\pdfoutput=1
\documentclass[10pt,twocolumn,letterpaper]{article}

\usepackage[pagenumbers]{cvpr} 

\definecolor{cvprblue}{rgb}{0.21,0.49,0.74}
\usepackage[unicode, bookmarks=true, bookmarksopen=true, pdfencoding=auto]{hyperref}
\hypersetup{pagebackref,breaklinks,colorlinks,allcolors=cvprblue}

\usepackage{url}
\usepackage{enumitem}
\usepackage{booktabs}
\usepackage{amsthm}
\usepackage{amsmath}
\usepackage{amssymb}
\usepackage{multirow}
\usepackage{longtable}
\usepackage{setspace}
\usepackage{nicefrac}
\usepackage{xspace}
\usepackage{graphicx}
\usepackage{color}
\usepackage{colortbl}
\usepackage{setspace}
\usepackage{dsfont}
\usepackage{soul}
\usepackage{longtable}
\usepackage{caption}
\usepackage{subcaption}
\usepackage{algpseudocode, algorithm}
\usepackage[T1]{fontenc}
\usepackage{wrapfig}
\usepackage{enumitem}
\usepackage{multirow}
\usepackage{color}
\usepackage[dvipsnames]{xcolor}

\usepackage{thmtools,thm-restate}

\theoremstyle{definition}

\newtheorem*{property*}{Property}

\usepackage{apptools}
\AtAppendix{\counterwithin{proposition}{section}}
\AtAppendix{\counterwithin{corollary}{section}}
\AtAppendix{\counterwithin{lemma}{section}}

\usepackage[capitalize]{cleveref}
\crefname{section}{Sec.}{Secs.}
\Crefname{section}{Section}{Sections}
\Crefname{table}{Table}{Tables}
\crefname{table}{Tab.}{Tabs.}

\def\paperID{4125} 
\def\confName{CVPR}
\def\confYear{2025}

\title{COUNTS: Benchmarking Object Detectors and  Multimodal Large Language Models under Distribution Shifts}

\author{Jiansheng Li$^{\dag}$, Xingxuan Zhang$^{\dag}$, Hao Zou, Yige Guo, Renzhe Xu, Yilong Liu \\
Chuzhao Zhu, Yue He, Peng Cui*\\
Department of Computer Science, Tsinghua University \\
{\tt\small{lijs23@mails.tsinghua.edu.cn,cuip@tsinghua.edu.cn}}
}

\begin{document}
\maketitle

\begin{abstract}
Current object detectors often suffer significant performance degradation in real-world applications when encountering distributional shifts. 
Consequently, the out-of-distribution (OOD) generalization capability of object detectors has garnered increasing attention from researchers. Despite this growing interest, there remains a lack of a large-scale, comprehensive dataset and evaluation benchmark with fine-grained annotations tailored to assess the OOD generalization on more intricate tasks like object detection and grounding.
To address this gap, we introduce COUNTS, a large-scale OOD dataset with object-level annotations. COUNTS encompasses 14 natural distributional shifts, over 222K samples, and more than  1,196K labeled bounding boxes. 
Leveraging COUNTS, we introduce two novel benchmarks: O(OD)$^2$ and OODG. O(OD)$^2$ is designed to comprehensively evaluate the OOD generalization capabilities of object detectors by utilizing controlled distribution shifts between training and testing data. OODG, on the other hand, aims to assess the OOD generalization of grounding abilities in multimodal large language models (MLLMs). 
Our findings reveal that, while large models and extensive pre-training data substantially enhance performance in in-distribution (IID) scenarios, significant limitations and opportunities for improvement persist in OOD contexts for both object detectors and MLLMs. In visual grounding tasks, even the advanced GPT-4o and Gemini-1.5 only achieve 56.7\% and 28.0\% accuracy, respectively. We hope COUNTS facilitates advancements in the development and assessment of robust object detectors and MLLMs capable of maintaining high performance under distributional shifts.


\end{abstract}
\footnotetext  [1]{$^{\dagger}$Equal Contribution, $^*$Corresponding Author}
\footnotetext  [2]{Code and Dataset are available at https://github.com/\\jiansheng-li/COUNTS\_benchmark.}  
\section{Introduction}

Extensive literature demonstrates that when the independent and identically distributed (IID) assumption between training and test data is violated, existing models may experience significant performance degradation~\citep{wang2022generalizing,zhou2022domain,koh2021wilds,hendrycks2021many}. Even current multimodal large language models (MLLMs)\citep{openai2023gpt4,team2023gemini,alayrac2022flamingo,gao2023llama,bai2023qwen} exhibit limited out-of-distribution generalization capabilities, severely hindering their usability and trustworthiness in specific domains~\citep{yang2023dawn,han2023well,zhang2024out}.

The challenges inherent in out-of-distribution (OOD) generalization have spurred significant research interest. However, existing efforts have predominantly centered on image classification tasks~\citep{li2017deeper,koh2021wilds,gulrajani2020search}, leaving a considerable gap in our understanding of how models perform on more intricate, real-world tasks. Furthermore, while recent works on MLLMs have aimed for comprehensive evaluation~\citep{yue2023mmmu,bai2023touchstone,lu2023mathvista}, they often neglect the crucial aspect of grounding model responses within the visual context. This grounding is indispensable for sophisticated applications such as detailed visual comprehension, interactive embodied agents, and localized content manipulation. Recent advancements allowing models to process user-specified regions via bounding boxes represent a promising direction~\citep{rasheed2023glamm}, but a thorough investigation of grounding abilities from an OOD generalization perspective remains an open research area. This underscores the need for novel benchmarks and datasets specifically designed to rigorously assess these critical facets of model performance.

This discrepancy can be attributed to the scarcity of suitable benchmark datasets. While comprehensive benchmarks for classification tasks abound~\citep{koh2021wilds,Peng19,hendrycks2019benchmarking,hendrycks2021natural}, there is a dearth of robust benchmarks offering finer-grained annotations. Previous studies have often relied on synthetically corrupted data for robustness evaluation~\citep{michaelis2019benchmarking}, but the extent to which such simulations accurately reflect real-world scenarios remains debatable. Consequently, other research has turned to collecting images from the internet to construct datasets. For example, road scene datasets~\citep{Yu_2020_CVPR,Cordts16,Johnson17,Geiger12} have been employed to benchmark the domain generalization of detectors~\citep{Khodabandeh19,zhenwei19,Chuang21,Zhu_2019_CVPR}. However, such scene-specific datasets lack generalizability and domain diversity, potentially leading to biased robustness assessments. Moreover, recent efforts have sought to analyze the out-of-distribution generalization capabilities of MLLMs~\citep{han2023well,zhang2024out}. Nonetheless, due to the absence of appropriate benchmarks, these studies have primarily focused on simple image-level classification tasks~\citep{jiang2024many}, neglecting the investigation of fine-grained visual grounding generalization in large models.

In this work, we introduce Common Objects UNder disTribution Shifts (COUNTS), a large-scale finely annotated dataset designed for OOD generalization research. COUNTS comprises 14 distinct domains, 35 categories, and 222,234 samples, all derived from real-world images and meticulously labeled through a combination of automated pipelines and human annotation. As the first real-world dataset to support both training and testing of OOD generalization for object detection and grounding tasks, COUNTS enables us to propose two novel benchmarks: OOD in Object Detection (O(OD)$^2$) for evaluating object detectors and OOD Grounding (OODG) for assessing the grounding generalization capabilities of MLLMs.

In OODG, given the lack of transparency regarding MLLMs' training data, we define distribution shift as the discrepancy between few-shot examples (in-context learning) and test samples. This definition reflects the real-world usage of MLLMs, where users typically provide prompts and examples before deployment, underscoring the importance of generalization based on given information.

We propose five settings in OODG:

\textbf{1. Zero-shot capability} assesses the MLLM's zero-shot grounding ability across diverse domains.

\textbf{2. ICL with IID samples} evaluates the MLLM's performance when the in-context examples (ICE) and test samples are independent and identically distributed.

\textbf{3. Generalization under covariate shifts} examines the MLLM's generalization ability when faced with unknown input sample distributions.

\textbf{4. Generalization under label shifts} investigates the MLLM's generalization ability when faced with unknown test label distributions.

\textbf{5. Generalization under spurious correlation shifts} assesses the MLLM's generalization ability on unknown test sample distributions when spurious correlations between visual features and labels exist in the ICE.

Our contributions are summarized as follows:

\begin{itemize}[leftmargin=0.6cm]

\item We introduce COUNTS, the first large-scale dataset with fine-grained annotations specifically designed to support both object detection and grounding tasks under natural distribution shifts.

\item We propose O(OD)$^2$, a benchmark for evaluating the OOD generalization capabilities of object detectors using COUNTS. Through comprehensive analysis, we identify common factors that contribute to model robustness, providing insights that can inform the development of more resilient detection algorithms.

\item We introduce OODG, a benchmark with five distinct settings designed to assess the OOD generalization of grounding abilities in MLLMs. Leveraging COUNTS, we investigate the impact of ICL distribution shifts on the grounding performance of MLLMs.
\end{itemize}

\section{Related Works}
In this section, we briefly review the previous works on object detection and multimodal large language models.

\textbf{Object Detection} There have been large amounts of literature proposing the models and benchmarks of object detectors. The models of object detectors can be categorized into two classes, that are single-stage detectors~\cite{liu2016ssd, lin2017focal, zhou2019objects, tan2020efficientdet, tian2022fully} and two-stage detectors~\cite{ren2015faster, girshick2014rich, girshick2015fast, lin2017feature, he2017mask}. They usually relies on the convolutional layer~\cite{lecun1998gradient} to extract the representation of images. Recently, motivated by the advancement of transformer architecture~\cite{vaswani2017attention}, many transformer-based object detectors~\cite{carion2020end, zhu2020deformable} are proposed. To evaluate the performance of the various object detector, several datasets are proposed, such as COCO~\cite{lin2014microsoft} and VOC~\cite{everingham2015pascal}. However, in the wild scenarios, the object detectors may face the impact of corruption and distribution shifts. To incorporate these factors into consideration, COCO-C~\cite{michaelis2019benchmarking} and COCO-O~\cite{mao2023coco} are sequentially proposed. COCO-C adds synthetic corruptions into the images of COCO to obtain the perturbated images. COCO-O aims to evaluate the out-of-distribution generalization of detectors and account for natural distribution shift.

\textbf{Multimodal Large Language Models} In recent years, many multimodal large language models are developed to jointly encode vision and language~\cite{li2020oscar,zhang2021vinvl,chen2020uniter,tan2019lxmert,radford2021learning}. They are pre-trained with massive amount of vision data and text on the web. As these multimodal large language models evolve, they are becoming increasingly excelling in understanding the content combining visual and textual information. Based on the pre-trained large models, instruction-tuning~\cite{awadalla2023openflamingo,li2023mimic,monajatipoor2023metavl,huang2023sparkles,zhao2023mmicl} can be applied and improve their capability of instruction following. 
Researchers can harness the power of multimodal large language models to conduct grounding~\cite{rasheed2023glamm} and provide insightful answers to questions pertaining to images, such as the object information in the images.

\textbf{Benchmark of MLLM} To assess the capabilities of multimodal large language models, the research community has developed massive amount of benchmarks. Some benchmarks focus on the single-task evaluation. For example, VQA~\cite{goyal2017making}, OK-VQA~\cite{marino2019ok}, and GQA~\cite{hudson2019gqa} are all dedicated to the pursuit of excellence in the domain of visual question answering. To 
compensate the insufficiency of these benchmarks in holistical assessment of multimodal large language models~\cite{liu2023mmbench,yu2023mm, xu2023lvlm, li2023seed, liu2023mmbench, yu2023mm}, numerous benchmarks are proposed to evaluate the other aspect of the models, including adversarial robustness~\cite{zhao2024evaluating}, hallucination~\cite{cui2023holistic}, mathematical ability~\cite{lu2023mathvista} and multimodality handling~\cite{mialon2023gaia,yue2023mmmu}. Although these benchmarks inspect the capabilities of multimodal large language models in different tasks, they neglect the impact of distribution shift, which is the focus of this paper.
\section{The COUNTS Dataset}
\subsection{Overview of COUNTS}
We introduce Common Objects UNder disTribution Shifts (COUNTS), a large-scale, fine-grained annotated dataset designed to facilitate the training and evaluation of traditional object detectors, as well as to assess the OOD grounding capabilities of MLLMs. COUNTS encompasses 35 object categories (visual concepts), 14 distinct domains, 222,234 images, and 1,196,114 annotated bounding boxes. All images in COUNTS are sourced from real-world scenes, ensuring the absence of artificially generated samples. As noted in prior work~\citep{mao2023coco,zhang2023nico++}, naturally occurring distribution shifts more accurately reflect the challenges encountered in real-world applications, while also presenting greater potential for meaningful solutions. Furthermore, each domain within COUNTS includes the full spectrum of object categories, guaranteeing the validity of cross-distribution testing.

\begin{figure*}[t]
    \centering
    \includegraphics[width=1.0\linewidth,bb=0 0 2445 742.5]{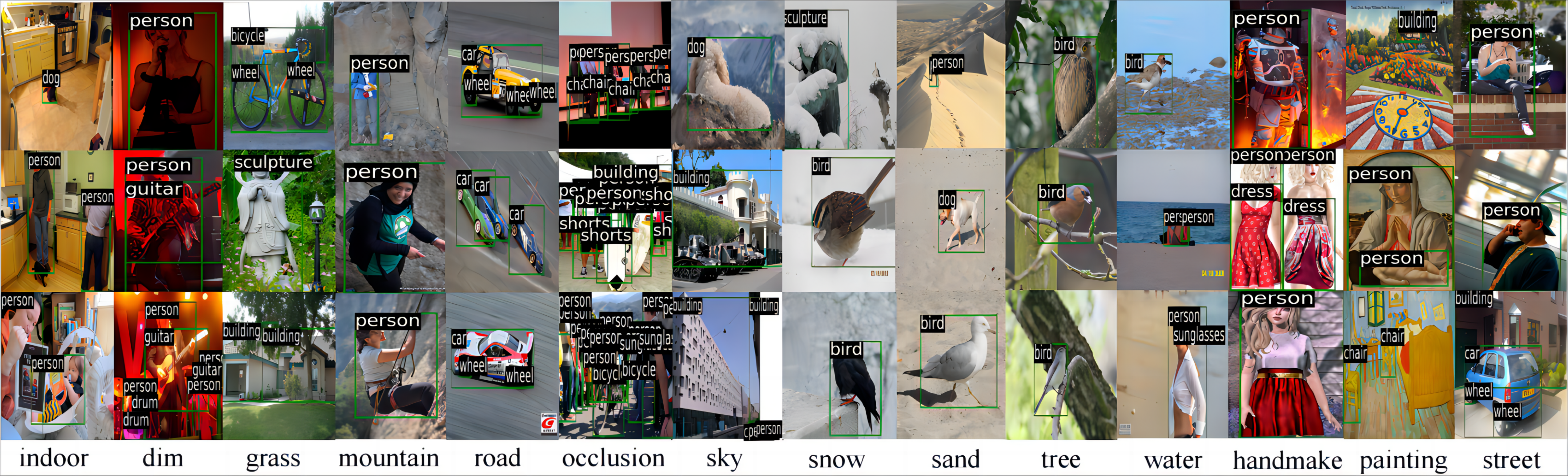}
    \caption{Examples of images in COUNTS. Each image is annotated with domain and objects.}
    \label{fig:showcase}

\end{figure*}

COUNTS provides significant distribution shifts to support the evaluation of OOD generalization abilities. Example images are shown in Figure \ref{fig:showcase}. Existing OOD generalization datasets primarily focus on classification tasks, limiting their ability to examine fine-grained spatial visual information. Different domains may differentially impact various regions within the same image (e.g., areas with varying snow thickness), posing greater challenges for models to perceive visual information at a spatially granular level. Additionally, COUNTS's ability to support the training of detectors, coupled with its abundance of domains and categories, allows for the creation of controlled distribution shifts by adjusting the training and testing distributions.

\subsection{domain Selection}
To ensure that the selected domains accurately reflect the data distributions likely encountered in real-world model deployment, we focused on domains that are prevalent in real-world scenarios and have the potential to significantly impact image pixel distributions, while also striving for mutual independence among the chosen domains.  Ultimately, we identified 14 domains: \textit{dim, painting, snow, sand, handmade, street, road, water, grass, indoor, mountain, sky, tree}, and \textit{occlusion}. These domains are frequently encountered in daily life and are likely to exhibit distinct associations with objects. It is important to note that objects within the samples are typically in the foreground, while domains often serve as background elements or overall image attributes. Therefore, we annotate domains at the image level, while bounding boxes are provided at the object level. A more comprehensive description and additional information regarding these domains can be found in Appendix A.

\subsection{Data Collection}
We primarily employed a three-stage data selection and annotation process. In the first stage, we conduct an initial screening of publicly available image datasets containing image captions, based on our predefined domain space. We consider Open Images~\citep{kuznetsova2020open}, Visual Genome (VG)~\citep{krishna2017visual}, RefCOCO~\citep{kazemzadeh2014referitgame}, RefCOCO+~\citep{kazemzadeh2014referitgame}, RefCOCOg~\citep{yu2016modeling}, Flickr30K~\citep{plummer2015flickr30k}, and GranD~\citep{rasheed2023glamm}. In total, more than 5 million images with domains are initially screened.
In the second stage, we engaged human annotators to verify and annotate the domain information of the selected images. Only samples that were consistently identified as belonging to the same domain by two independent annotators were retained and the labeled domain is considered as the ground-truth domain for each image. 
In the third stage, we randomly sampled 10\% of the data within each domain to form the test and validation sets. We observe that datasets such as Open Images and GranD contained automatically annotated data, resulting in incomplete or inaccurate bounding box annotations for some objects. Our experiments reveal that training detectors on data with incomplete bounding box annotations did not significantly impact their performance. Conversely, incomplete or inaccurate annotations in the test or validation sets notably affected the reliability of the test results. Therefore, we employ human annotators to meticulously re-annotate the bounding boxes in the test and validation sets, ensuring that each object is labeled and verified by two independent annotators. A total of 23,000 images are re-annotated with bounding boxes by annotators. Please see more details in Appendix A.

\subsection{Comparison with Existing Benchmarks}

To further distinguish COUNTS from existing OOD generalization benchmarks, Table \ref{tab:datasets} provides a comparative overview of relevant dataset characteristics. While current large-scale OOD benchmarks predominantly focus on image classification, those offering finer-grained annotations (e.g., object detection) tend to be limited in scope. Existing datasets often lack either a sufficient diversity of domains to thoroughly assess OOD generalization capabilities or the scale necessary to support robust detector training and controlled experimentation with distribution shifts. COUNTS addresses these limitations by offering a substantial number of diverse domains, a vast corpus of natural images, and high-quality fine-grained annotations. This unique combination enables researchers to explore a wider range of advanced and complex OOD generalization tasks within the visual domain, including those requiring precise object localization and detection under varying conditions.

\begin{table*}[th]
\centering

\resizebox{0.75\linewidth}{!}{
\begin{tabular}{lccccc}
\toprule

Datasets & Task & Domain Num. & Class Num. & Natural Images & Images / Domain\\
PACS\citep{LiDa17} & \multirow{4}*{Classification} & 4 & 7 & 9,991 & 2,497 \\ 
VLCS\citep{FangChen13} &  & 4 & 5 & 10,729 & 2,682 \\
Officehome\citep{Venkateswara17} & & 4 & 65 & 15,588 & 3,897 \\
DomainNet\citep{Peng19} &  & \textbf{6} & \textbf{345} & \textbf{586,575} & \textbf{97,762} \\
\cmidrule(lr){1-6}
OOD-CV\citep{Zhao22} & & 5 & 10 & 2,632 & 526 \\
Comic2k\citep{Inoue_2018_CVPR} & & 1 & 6 & 2,000 & 2,000 \\
Clipart\citep{Inoue_2018_CVPR}  & Object & 1 & 20 & 1,000 & 1,000 \\
Watercolor\citep{Inoue_2018_CVPR} & Detection\& & 1 & 6 & 2,000 & 2,000 \\
COCO-C\citep{Claudio19} & Grounding & \textbf{15} & \textbf{80} & 0 & 0 \\
COCO-O\citep{Xiaofeng23} & & 6 & \textbf{80} & 6,782 & 1,130 \\
COUNTS (ours) & & 14 & 35 & \textbf{222,234} & \textbf{15,874}  \\

\bottomrule
\end{tabular}}
\caption{Overview of current OOD generalization and robust detection benchmarks.}
\label{tab:datasets}
\end{table*}
\section{Benchmarks}

\subsection{OOD in Object Detection (O(OD)$^2$)}
O(OD)$^2$ is designed to assess the OOD generalization capabilities of existing object detectors and to investigate the key factors that influence their generalization performance, thereby providing insights for future advancements.

Given the strong performance of current detectors under IID conditions, we specifically focus on their performance in unknown domains where the test domains do not overlap with the training domains. As COUNTS maintains a consistent category space across all domains, we utilize the full set of categories for both training and testing phases.

Furthermore, prior research suggests that evaluating models on multiple target distributions, rather than a single one, more accurately reflects real-world scenarios with numerous potential unknown distributions and mitigates the risk of unfairness caused by target distribution information leakage~\citep{zhang2023nico++,yu2023rethinking}. Consequently, we employ multiple domains as target domains, with the remaining domains serving as the training set. Specifically, in the main paper, we adopt \textit{sky, occlusion, grass, water, dim}, and \textit{handmake} as the target domains due to their representation of diverse distribution shift patterns. For example, the \textit{sky} domain presents challenges due to the high variability in angle and lighting conditions. The \textit{occlusion} domain tests the model's robustness to partial or complete object obstruction, while the \textit{grass} and \textit{water} domains introduce complex backgrounds with varying textures and reflections. The \textit{dim} domain evaluates performance under low-light conditions, and the \textit{handmake} domain assesses generalization to artistic renditions or sketches of objects.

\subsection{OOD in Grounding (OODG)}
With OODG, we investigate the generalization abilities of MLLMs in the context of grounding tasks, a critical component for various real-world applications such as autonomous driving and visual perception in autonomous systems.

Currently, there is no universally accepted definition for OOD generalization in MLLMs. Traditional OOD generalization approaches rely on artificially constructed distribution shifts between training and test data. However, due to the unknown nature of current MLLMs' pre-training and fine-tuning data, it is challenging to characterize the distribution shifts between their training and test data.

Given the limited information available regarding the pre-training and fine-tuning data of many MLLMs, we propose to define distribution shifts within the ICL stage. This approach allows us to examine the generalization capabilities of MLLMs when there are distribution shifts between the ICE and the actual test samples. This setting is well-justified, as MLLM parameters remain fixed in many applications, and defining distribution shifts during ICL enables assessment of their generalization to diverse scenarios in real-world deployment. It is worth noting that COUNTS contains a sufficient amount of high-quality data to support fine-tuning the grounding capabilities of MLLMs. This opens up the possibility of investigating the models' generalization abilities when distribution shifts occur between the fine-tuning data and test data. However, this is not the primary focus of this work.

We utilize data from the human-annotated test and validation sets to construct the evaluation framework for this problem. We primarily consider three testing scenarios:

\textbf{1. Visual Grounding}. Given an image and a region of interest, we query the multimodal large language model and assess the accuracy of its textual response.

\textbf{2. Recognition and Localization}. Given an image and a question, we evaluate the accuracy of the model's predicted bounding box coordinates for the relevant region.

\textbf{3. Visual and Semantic Mapping}. Given an image and a sequence of textual descriptions, we assess the accuracy of the model's ability to map the descriptions to the corresponding regions in the image.

For generalization evaluation, we consider five distinct settings:

\textbf{1. Zero-shot Generalization}.  This setting investigates the ability of MLLMs to recognize visual concepts across various domains without any prior examples.
\textbf{2. I.I.D. ICL}.  This setting examines the i.i.d. generalization capability of MLLMs when provided with examples drawn from the same distribution as the test data.
\textbf{3. ICL with Covariate Shifts}. This setting explores the impact of covariate shifts between in-context examples and test samples on model performance. It aligns with existing OOD generalization research, where examples and test samples originate from different domains, to assess the model's ability to generalize to unseen sample distributions.
\textbf{4. ICL with Label Shifts}. This setting investigates the generalization capability of MLLMs when label shifts exist between in-context examples and test samples. It aims to determine whether the distribution of labels in the examples significantly affects model performance when the distribution of visual concepts in the test data is unknown.
\textbf{5. ICL with Spurious Correlation Shifts}. This setting examines whether MLLMs are misled by spurious correlations between image features and descriptions in the in-context examples, causing them to overlook more fundamental relationships. Specifically, due to sampling bias, providing ICL samples may introduce statistical associations between irrelevant features and descriptions. For instance, if the concept \textit{cat} frequently appears in darker indoor settings while \textit{dog} appears predominantly outdoors in the ICL samples, a spurious correlation may arise between dim lighting and the concept \textit{cat} or between outdoor environments and the concept \textit{dog}. This could lead to erroneous predictions when the model encounters a cat in an outdoor setting.

\section{Experiments}
\label{sect:exp}

\subsection{OOD in Object Detection}
For O(OD)$^2$, we employ representative object detectors spanning two distinct architectures: the two-stage Faster R-CNN~\citep{ren2015faster} and the one-stage detectors RetinaNet~\citep{lin2017focal} and YOLOv9~\citep{wang2024yolov9}. All experiments were conducted using the MMDetection framework, adhering to the optimal training configurations specified in the respective original publications. The computational resources for this benchmark consisted of NVIDIA RTX 4090 GPUs. 
We investigate the impact of various advanced model architectures on the OOD generalization capabilities of object detectors, focusing on the backbone, neck, and head components. 
All experiments investigating model structures begin by training detectors from initial parameters pre-trained on ImageNet, ensuring a fair comparison across architectures. To further explore the influence of pre-training data on a detector's OOD generalization ability, we utilize diverse pre-training datasets for model initialization on representative model structures. Introduction of the evaluated models and more experimental setups such as training hyperparameter selection are detailed in Appendix B.

\begin{table*}[th]
\centering

\resizebox{0.95\linewidth}{!}{
\begin{tabular}{lccc|c|cccccc|c}
\toprule
& & & & VAL & \multicolumn{7}{c}{Test mAP}  \\ 
& & &  & mAP & Sky & Occlusion & Grass & Water & Dim & Handmake &Avg  \\ 
\midrule

\multirow{13}*{\rotatebox{90}{\textbf{Two-stage Detector}}} & \multirow{13}*{\textbf{Faster R-CNN}} & \multirow{5}*{\textbf{Backbone}} 
& RN-50~\citep{he2016deep} & 0.279 &0.178 &0.139&  0.156 &0.103 &0.140 &0.134 &0.135 \\
& & & RN-101~\citep{he2016deep}   & \textbf{0.325} &\textbf{0.212} &0.143&  0.166 &\textbf{0.110} &0.152 &\textbf{0.144} &0.148  \\
& & & RX-101-32x4d~\citep{xie2017aggregated}    &  0.310 &0.200& \textbf{0.163} & \textbf{0.179}&0.109 & \textbf{0.159} & 0.132 & \textbf{0.150} \\
& & & Swin-T ~\citep{liu2021swin}  &  0.308 &0.192& 0.158 & 0.163&0.102 & 0.155& 0.129 & 0.142 \\
& & & PVTv2-B2  ~\citep {wang2022pvtv2} &  0.303 &0.189& 0.156 & 0.161&0.103 & 0.153& 0.126 & 0.141  \\
\cmidrule(lr){3-12}

& & \multirow{3}*{\textbf{Neck}} 
& FPN~\citep{lin2017feature}   & 0.279 &0.178 &0.139&  0.156 &0.103 &\textbf{0.140} &\textbf{0.134} &0.135 \\
& & & PAFPN~\citep{liu2018path}  & \textbf{0.309} &0.184 & \textbf{0.148}& \textbf{0.158} & 0.140& 0.109& 0.119 &\textbf{0.140} \\
& & & NAS-FPN~\citep{ghiasi2019fpn}   & 0.298 &\textbf{0.186} & 0.141& 0.153 & \textbf{0.143}& 0.108& 0.113 &0.136 \\
\cmidrule(lr){3-12}

& & \multirow{5}*{\textbf{Head}} 
& Standard   &  0.279 &0.178 &0.139&  0.156 &0.103 &0.140 &0.134 &0.135 \\
& & & Cascade ~\citep{cai2019cascade}  &0.339 &0.193& 0.162& 0.172& 0.111 &0.154& 0.146 & 0.149 \\
& & & SABL~\citep{wang2020side}     &\textbf{0.347}&\textbf{0.218}& \textbf{0.177}& \textbf{0.182}& 0.135 &\textbf{0.165}& \textbf{0.158} & \textbf{0.162} \\
& & & 2\_Heads~\citep{wu2020rethinking}   &0.308 &0.197& 0.163 & 0.163 &0.120 &0.150 &0.133&0.149 \\
& & & Groie~\citep{rossi2021novel}   &0.309 &0.188&0.153 & 0.164 &\textbf{0.148} &0.150 &0.131&0.141\\

\midrule

\multirow{12}*{\rotatebox{90}{\textbf{One-stage Detector}}} & \multirow{12}*{\textbf{RetinaNet}} & \multirow{5}*{\textbf{Backbone}} 
& RN-50~\citep{he2016deep}  &0.331& 0.199 &0.189 &  0.166 &0.108 &0.179 &\textbf{0.149} &0.156  \\
& & & RN-101 ~\citep{he2016deep}     &0.342& \textbf{0.203} &\textbf{0.194} &  0.175 &0.110 &\textbf{0.192} &\textbf{0.149} &\textbf{0.162} \\
& & & RX-101-32x4d~\citep{xie2017aggregated}  & 0.302 &0.189& 0.158 & 0.169&0.104 & 0.147 & 0.131 & 0.144 \\
& & & Swin-T ~\citep{liu2021swin}   &\textbf{0.340}& 0.201 &0.191 &  \textbf{0.179} &\textbf{0.117} &0.188 &0.146 &\textbf{0.162} \\
& & & PVTv2-B2 ~\citep {wang2022pvtv2}  &0.338& 0.192 &0.184 &  0.175 &0.109 &0.183 &0.143 &0.158 \\
\cmidrule(lr){3-12}

& & \multirow{3}*{\textbf{Neck}} 
& FPN~\citep{lin2017feature}  &0.331& 0.199 &\textbf{0.189} &  \textbf{0.166} &0.108 &\textbf{0.179} &\textbf{0.149} &0.156\\
& & & PAFPN~\citep{liu2018path}   &\textbf{0.339}&\textbf{0.205}& 0.178& \textbf{0.166}& \textbf{0.123} &0.162& 0.138 & \textbf{0.157} \\
& & & NAS-FPN~\citep{ghiasi2019fpn}   &0.306&0.146& 0.141& 0.157& 0.113 &0.128& 0.127 & 0.133 \\
\cmidrule(lr){3-12}

& & \multirow{4}*{\textbf{Head}} 
& Standard   &0.331& 0.199 &\textbf{0.189} &  0.166 &0.108 &\textbf{0.179} &0.149 &0.156 \\
& & & SABL~\citep{wang2020side}  &\textbf{0.345}&\textbf{0.214}& \textbf{0.189}& \textbf{0.180}& \textbf{0.141} &0.168& \textbf{0.156} & \textbf{0.161} \\
& & & FSAF~\citep{zhu2019feature}  &0.332&0.174& 0.172& 0.168& 0.123 &0.162& 0.127 & 0.153 \\
& & & FreeAnchor~\citep {zhang2019freeanchor} &0.298&0.166& 0.115& 0.163& 0.111 &0.138& 0.116 & 0.131 \\




\midrule

\multicolumn{3}{c}{\textbf{YOLO V9}~\citep{wang2024yolov9}} & &0.282&0.190& 0.118& 0.160& 0.120 &0.153& 0.121 & 0.150 \\
\multicolumn{3}{c}{\textbf{DETR}~\citep{DBLP:journals/corr/abs-2005-12872}}& & 0.348& 0.211 & 0.183 &0.185& 0.121 & 0.209 & 0.159& 0.168  \\
\multicolumn{3}{c}{\textbf{DINO}~\citep{DBLP:journals/corr/abs-2104-14294}} & & 0.384& \textbf{0.250} & \textbf{0.250} &0.229& \textbf{0.169} & 0.212 & \textbf{0.181}& \textbf{0.213}  \\
\multicolumn{3}{c}{\textbf{DINO V2} ~\citep{oquab2023dinov2}}& &\textbf{ 0.389}& 0.252 & 0.249 &\textbf{0.231}& 0.165 & \textbf{0.213} & \textbf{0.181}& \textbf{0.213}  \\

\bottomrule
\end{tabular}}
\caption{Comparison of object detectors with different backbone, neck, and head. The best results of components for each detector are highlighted in bold font.}
\label{tab:oodod}
\end{table*}

The main results of O(OD)$^2$ with various detector architectures and models are shown in Table \ref{tab:oodod}. 
Existing detectors exhibit limited OOD generalization capabilities and improvements in IID performance do not necessarily translate to enhanced OOD generalization. Performance across various OOD scenarios was notably suboptimal compared with their performance on IID benchmarks. In some cases, IID improvements may even negatively impact OOD performance. For example, with stronger neck NAS-FPN and head FreeAnchor, the detectors show weaker generalization across domains.

For two-stage detectors, modifications to the head architecture yield more significant improvements than strengthening the backbone. For one-stage detectors, both stronger backbones and heads contribute to improved results. This suggests that the head can be the key to learning strong local representations and reducing environmental impacts. Thus optimizing the head design may be a more effective strategy for enhancing OOD generalization. Moreover, improvements in the neck architecture have limited impacts on OOD generalization. 
Among all evaluated models, DINO and DINOv2 demonstrated the most favorable performance, significantly outperforming other detectors. This suggests that, in addition to model architecture, training strategies can exert a substantial influence on a model's OOD generalization capabilities.

\begin{table*}[th]
\centering

\resizebox{\linewidth}{!}{
\begin{tabular}{lcccccccccc|c}
\toprule
& & Backbone method & Pretraining Data &  VAL & \multicolumn{7}{c}{mAP on COUNTS}  \\ 
& & &  & mAP & Sky & Occlusion & Grass & Water & Dim & Handmake & Avg  \\ 
\midrule

\multirow{5}*{\rotatebox{90}{\textbf{Two-stage}}} & \multirow{5}*{\textbf{Faster R-CNN}} &   Supervised  & IN-1k & 0.301 &0.193 & 0.140 &0.153 &0.112 &0.148 &0.136 & 0.141 \\
 & & Sup\_timm~\citep{wightman2021resnet}    & IN-1k&\textbf{ 0.323} &\textbf{0.210} &\textbf{ 0.147} &\textbf{0.161} &\textbf{0.118} &\textbf{0.165} &\textbf{0.139} &\textbf{ 0.167}\\
 & & MoCov2~\citep{chen2020improved}     & IN-1k &0.298 &0.193 & 0.136 &0.148 &0.114 &0.139 &0.118 & 0.136 \\
 & & Supervised    & IN-21k~\citep{ridnik2021imagenet} &0.313 &0.201 & 0.142 &0.158 &0.114 &0.152 &0.134 & 0.146 \\

\cmidrule(lr){2-12}

\multirow{5}*{\rotatebox{90}{\textbf{One-stage}}} & \multirow{5}*{\textbf{RetinaNet}} & 
Supervised  & IN-1k & 0.309 &0.197 & 0.143 &0.156 &0.113 &0.151 &0.139 & 0.145 \\
 & & Sup\_timm~\citep{wightman2021resnet}     & IN-1k &\textbf{ 0.325} &\textbf{0.214} & \textbf{0.149} &\textbf{0.162} &\textbf{0.123 }&\textbf{0.170} &\textbf{0.142} & \textbf{0.170} \\
 & & MoCov2~\citep{chen2020improved}    & IN-1k &0.303 &0.189 & 0.131 &0.144 &0.112 &0.135 &0.116 & 0.133  \\
 & & Supervised    & IN-21k~\citep{ridnik2021imagenet} &0.313 &0.201 & 0.142 &0.158 &0.114 &0.152 &0.134 & 0.146\\


\bottomrule
\end{tabular}}
\caption{Comparison of object detectors with different backbone and pretraining methods. }
\label{tab:pretrain}
\end{table*}

In Table \ref{tab:pretrain}, we analyze the impact of pre-training data and methods on the OOD generalization capabilities of object detectors. Our findings reveal that increasing the size of pre-training data or employing self-supervised pre-training methods does not yield significant improvements in generalization. However, utilizing advanced training methodologies can substantially enhance model performance. For instance, Sup\_timm~\citep{wightman2021resnet}, when trained on the same data, achieves a 2.67\% absolute improvement in mean average precision (mAP) and an 18.9\% relative improvement compared to traditional ImageNet pre-training.

\begin{table*}[th]
\centering
\resizebox{\linewidth}{!}{
\begin{tabular}{lccccccccccccccccc|c}
\toprule
\multirow{2}*{Setting} & \multirow{2}*{Model} & \multirow{2}*{ICE}  & \multicolumn{3}{c}{Sky} & \multicolumn{3}{c}{Occlusion} &  \multicolumn{3}{c}{Water} & \multicolumn{3}{c}{Dim}&\multicolumn{3}{c}{Handmake} & \multirow{2}*{Overall} \\ 
& & & S & M & L & S & M & L & S & M & L & S & M & L & S & M & L & \\

\midrule

\multirow{4}*{1} & GPT-4o &  0 &0.434&\textbf{0.727} &0.889 &0.362 &\textbf{0.891} &\textbf{0.805}&\textbf{0.554} &0.797 &0.860 &\textbf{0.626} & 0.727& 0.737&\textbf{0.353} &0.595 &0.535&\textbf{0.659} \\ 
\cmidrule(lr){2-19}
 & Gemini-1.5-Flash &  0 &\textbf{0.444} &0.667 & 0.778& 0.355&0.674 &0.628 &0.533 &0.783 &0.858&0.347 & 0.716& 0.570&0.333 &\textbf{0.608} &\textbf{0.575}&0.591 \\ 
  \cmidrule(lr){2-19}
&GLaMM~\cite{rasheed2024glamm}
&0&0.333&0.667&\textbf{1.000}&\textbf{0.387}&0.714&0.444&\textbf{0.554}&\textbf{0.857}&\textbf{1.000}&0.429&\textbf{0.857}&\textbf{0.857}&0.333&0.429&0.513&0.625\\
 \cmidrule(lr){2-19}
&Qwen2-VL~\cite{wang2024qwen2}
&0&0.286&0.667&0.889&0.333&0.667&0.355&0.533&0.667&0.750&0.385&0.625&0.714&0.286&0.444&0.555&0.544\\
\midrule

\multirow{4}*{2} & \multirow{2}*{GPT-4o} &  1&\textbf{0.445}&0.778&0.778&\textbf{0.571}&\textbf{1.000}&\textbf{1.000}&0.143&0.429&\textbf{1.000}&\textbf{0.333}&0.771&0.750&0.000&0.680&\textbf{1.000}&0.646 \\ 
&  & 2 &\textbf{0.445}&0.667&0.778&0.143&\textbf{0.857}&\textbf{1.000}&\textbf{0.554}&0.797&0.860&\textbf{0.333}&0.778&0.778&0.000&\textbf{0.825}&\textbf{1.000}&0.654 \\
\cmidrule(lr){2-19}
 & \multirow{2}*{Gemini-1.5-Flash} &  1 &0.444&0.714&\textbf{1.000}&0.143&\textbf{1.000}&0.857&0.286&\textbf{0.857}&\textbf{1.000}&0.222&\textbf{1.000}&\textbf{0.889}&0.000&0.637&\textbf{1.000}&0.670 \\ 
&  & 2 &0.444&\textbf{0.857}&\textbf{1.000}&0.143&\textbf{1.000}&\textbf{1.000}&0.143&\textbf{0.857}&\textbf{1.000}&\textbf{0.333}&\textbf{1.000}&\textbf{0.889}&0.000&0.671&\textbf{1.000}&\textbf{0.689} \\

\midrule

\multirow{4}*{3} & \multirow{2}*{GPT-4o} &  1 &0.143&0.357&\textbf{0.857}&0.357&0.786&\textbf{0.929}&0.143&0.429&0.786&0.222&0.722&0.667&0.000&\textbf{0.889}&\textbf{1.000}&0.552 \\ 
&  & 2 &\textbf{0.429}&\textbf{0.556}&0.492&\textbf{0.385}&\textbf{1.000}&\textbf{0.929}&0.143&0.429&\textbf{0.929}&\textbf{0.278}&\textbf{0.778}&\textbf{0.778}&0.000&0.714&\textbf{1.000}&\textbf{0.589} \\
\cmidrule(lr){2-19}
 & \multirow{2}*{Gemini-1.5-Flash} &  1 &0.222&0.143&0.143&0.143&0.143&0.143&\textbf{0.286}&\textbf{0.714}&0.857&0.111&0.444&0.444&0.000&0.571&\textbf{1.000}&0.358 \\ 
&  & 2 &0.000&0.000&0.222&0.143&0.143&0.143&\textbf{0.286}&\textbf{0.714}&0.857&0.222&0.444&0.444&0.000&0.571&\textbf{1.000}&0.346 \\

\midrule

\multirow{4}*{4} & \multirow{2}*{GPT-4o} &  1 & 0.133&\textbf{0.286}&\textbf{1.000}&0.429&\textbf{0.857}&\textbf{0.857}&\textbf{0.286}&0.429&\textbf{0.857}&0.000&0.556&0.667&\textbf{0.342}&0.432&\textbf{1.000}&\textbf{0.533} \\ 
&  & 2 &\textbf{0.143}&\textbf{0.286}&\textbf{1.000}&0.286&0.571&\textbf{0.857}&\textbf{0.286}&0.429&\textbf{0.857}&\textbf{0.111}&\textbf{0.667}&\textbf{0.778}&0.000&0.424&\textbf{1.000}&0.503\\
\cmidrule(lr){2-19}
 & \multirow{2}*{Gemini-1.5-Flash} &  1 & 0.000&0.111&0.143&0.111&0.133&0.127&\textbf{0.286}&0.857&\textbf{0.857}&\textbf{0.111}&0.222&0.222&0.000&\textbf{0.714}&\textbf{1.000}&0.317\\ 
&  & 2 &0.111&0.111&0.143&0.111&0.137&0.111&\textbf{0.286}&\textbf{1.000}&0.714&0.000&0.222&0.222&0.000&0.424&\textbf{1.000}&0.297\\

\midrule

\multirow{4}*{5} & \multirow{2}*{GPT-4o} &  4 &0.286&0.500&\textbf{0.917}&0.308&0.444&\textbf{0.833}&\textbf{0.375}&0.556&\textbf{0.750}&0.333&0.588&0.769&0.133&0.352&0.462&0.507\\ 
&  & 8 &\textbf{0.714}&0.875&0.833&0.311&\textbf{0.556}&\textbf{0.833}&0.313&\textbf{0.778}&\textbf{0.750}&0.143&\textbf{0.800}&\textbf{0.923}&0.266&\textbf{0.412}&0.625&\textbf{0.609} \\
\cmidrule(lr){2-19}
 & \multirow{2}*{Gemini-1.5-Flash} &  4 &0.500&0.875&0.888&0.534&0.473&0.500&\textbf{0.375}&0.556&\textbf{0.750}&\textbf{0.347}&0.716&0.570&0.400&0.313&\textbf{0.769}&0.571\\ 
&  & 8 &0.467&\textbf{0.889}&0.875&\textbf{0.690}&0.544&0.812&0.313&\textbf{0.778}&\textbf{0.750}&\textbf{0.347}&0.716&0.570&\textbf{0.467}&0.353&0.462&0.602\\

\bottomrule
\end{tabular}}
\caption{Results of Visual Grounding. The selection accuracy of MLLMs is reported. \textit{S}, \textit{M}, and \textit{L} indicate small, medium, and large objects, respectively.}
\label{tab:grounding}
\end{table*}

\subsection{OOD in Grounding}

\begin{table*}[th]
\centering

\resizebox{0.8\linewidth}{!}{
\begin{tabular}{lccccccccccc|c}
\toprule
\multirow{2}*{Setting} & \multirow{2}*{Model} & \multirow{2}*{ICE}  & \multicolumn{3}{c}{Occlusion} & \multicolumn{3}{c}{Dim} & \multicolumn{3}{c}{Handmake} & \multirow{2}*{Overall} \\ 
& & & S & M & L & S & M & L & S & M & L   \\
\midrule
\multirow{2}*{1} & GPT-4o &  0 &\textbf{0.108} &\textbf{0.099} &\textbf{0.112}&\textbf{0.108} &\textbf{0.099} &\textbf{0.112}&\textbf{0.108} &\textbf{0.099} &\textbf{0.112}&\textbf{0.106} \\ 
\cmidrule(lr){2-12}
 & Gemini-1.5-Flash &  0 &0.083 &0.091 &0.090&0.101&\textbf{0.099}&0.083&0.100&0.002&0.089&0.082 \\ 
\midrule

\multirow{4}*{2} & \multirow{2}*{GPT-4o} &  1 &\textbf{0.113} &\textbf{0.068} &\textbf{0.101}&0.079 &0.101 &0.086&\textbf{0.099} &0.071 &0.100&\textbf{0.091} \\ 
&  & 2 &0.108 &0.059 &0.099&\textbf{0.083} &0.098 &0.083&\textbf{0.099} &0.068 &\textbf{0.101}&0.089 \\
\cmidrule(lr){2-12}
 & \multirow{2}*{Gemini-1.5-Flash} &  1 &0.107 &0.058 &\textbf{0.099}&0.068 &\textbf{0.106} &\textbf{0.097}&0.068 &0.054 &0.093&0.083 \\ 
&  & 2 &0.101 &0.048 &0.088&0.079 &0.101&0.086&\textbf{0.099} &\textbf{0.079} &0.081&0.085 \\

\midrule

\multirow{4}*{3} & \multirow{2}*{GPT-4o} &  1 &\textbf{0.041} &\textbf{0.035} &0.016&\textbf{0.008} &0.001&\textbf{0.072}&0.000 &\textbf{0.019} &\textbf{0.053}&\textbf{0.027} \\ 
&  & 2 &0.040 &0.015 &\textbf{0.032}&0.006 &\textbf{0.004}&0.065&0.000 &0.011 &0.038&0.023 \\
\cmidrule(lr){2-12}
 & \multirow{2}*{Gemini-1.5-Flash} &  1 &0.000 &0.000 &0.000&0.000 &0.000&0.000&0.000 &0.000 &0.000&0.000 \\ 
&  & 2 &0.000 &0.000 &0.000&0.000 &0.000&0.000&0.000 &0.000 &0.000&0.000 \\

\midrule

\multirow{4}*{4} & \multirow{2}*{GPT-4o} &  1 &\textbf{0.036} &0.020 &\textbf{0.101}&0.009 &0.014&0.003&0.100 &0.018 &\textbf{0.101}&\textbf{0.045} \\ 
&  & 2 &0.029 &0.018 &0.029&\textbf{0.019} &0.010&0.013&0.049 &\textbf{0.058} &0.017&0.027 \\
\cmidrule(lr){2-12}
 & \multirow{2}*{Gemini-1.5-Flash} &  1 &0.008 &0.074 &0.036&\textbf{0.019} &0.006&\textbf{0.018}&0.071 &0.028 &0.011&0.030 \\ 
&  & 2 &0.002 &\textbf{0.101} &0.019&0.008 &\textbf{0.038}&0.009&\textbf{0.101} &0.020 &0.099&0.044 \\

\midrule

\multirow{4}*{5} & \multirow{2}*{GPT-4o} &  4 &\textbf{0.019} &0.028 &0.013&\textbf{0.025} &\textbf{0.010}&0.009&0.053 &\textbf{0.091} &0.008&0.028 \\ 
&  & 8 &0.002 &\textbf{0.104} &\textbf{0.016}&0.018 &\textbf{0.010}&\textbf{0.029}&\textbf{0.084} &0.016 &\textbf{0.009}&\textbf{0.032} \\
\cmidrule(lr){2-12}
 & \multirow{2}*{Gemini-1.5-Flash} &  4 &0.000 &0.000 &0.000&0.000 &0.000&0.000&0.000 &0.000 &0.000&0.000 \\ 
&  & 8 &0.000 &0.000 &0.000&0.000 &0.000&0.000&0.000 &0.000 &0.000&0.000 \\

\bottomrule
\end{tabular}}
\caption{Results of Recognition and Localization. mAP is reported. }
\vspace{-10pt}

\label{tab:bbox}
\end{table*}

For OODG, we evaluated state-of-the-art multimodal large models including GPT-4, Gemini 1.5, and Claude 3. Due to space constraints, we focus on Visual Grounding and Recognition and Localization and primarily present results for GPT-4 and Gemini 1.5 Flash in the main body of this paper, with additional findings provided in Appendix C. 


\begin{itemize}[leftmargin=0.6cm]

    \item Visual Grounding.
In order to study the ability of multimodal large models to perform grounding within given regions, we implemented a methodology involving the use of highlighted bounding boxes to delineate objects or backgrounds within images. We then assessed the accuracy of the models' returned visual concepts. To ensure a thorough evaluation of visual recognition across a wide range of region types, we included all categories and domains as potential visual concepts, reporting accuracy rates for small (<32$\times$32), medium (from 32$\times$32 to 96$\times$96), and large (>96$\times$96) bounding boxes. A multiple-choice approach was employed, wherein all possible answers were presented in the prompt, and the accuracy of the model's selected visual concept was verified. Further details regarding prompt design and experimental specifics can be found in Appendix B.

    \item Recognition and Localization.
To assess the capability of MLLMs to perceive and comprehend comprehensive visual information within images, we designed a task wherein the MLLM is prompted to return the bounding box coordinates of specified visual elements. Input images were uniformly resized to a standardized resolution (448$\times$448) to mitigate potential errors arising from variations in image resolution and dimensions. We report the model's performance across all categories and domains, as well as its mean average precision (mAP) on small, medium, and large bounding boxes.
\end{itemize}

\paragraph{Zero-shot Generalization.} We show the main results of Visual Grounding and Recognition and Localization in Table \ref{tab:grounding}  and \ref{tab:bbox}, respectively. 
While current MLLMs demonstrate strong performance on image-level tasks, their performance on fine-grained grounding tasks remains suboptimal. Notably, both GPT-4o and Gemini-1.5 exhibit limitations when dealing with objects characterized by very small or very large bounding boxes, indicating a need for further enhancement of their localized perception capabilities. For instance, in the visual grounding task, GPT and Gemini achieve recognition accuracies of 35.3\% and 33.3\%, respectively, on small-sized bounding boxes within the \textit{handmake} domain, markedly lower than other results. In the Recognition and Localization task, both GPT-4o and Gemini-1.5-Flash show weak generalization and are still quite far from being usable.


\paragraph{ICL with IID Samples.} Our findings reveal that incorporating ICE drawn from the same distribution as the test samples significantly enhances Gemini's performance on visual grounding tasks. This improvement becomes more pronounced with increasing ICE quantity (see Appendix C for results with varying ICE counts). Conversely, GPT does not exhibit a similar enhancement.

\paragraph{ICL under Covariate / Concept / Spurious Correlation Shifts.}
Introducing distribution shifts, such as covariate and label shifts, into the ICE leads to a notable degradation in Gemini's performance compared to zero-shot learning. In the Visual Grounding task, average accuracy drops from a maximum of 57.0\% to 28.0\%, representing a relative decrease of 50.88\%. While GPT-4o also experiences a performance decline, it is less severe, with a maximum decrease of 12.1\%.

These results highlight two critical conclusions:

1. \textbf{Defining distribution shifts within the ICL phase is both valid and impactful.} This aligns with typical MLLM usage scenarios and is supported by our empirical findings, which demonstrate the substantial influence of ICL distribution shifts on model performance. Notably, covariate shift, a non-adversarial and naturally occurring distribution shift, closely approximates real-world applications~\citep{zhang2023nico++}. The presence of such shifts within ICL can lead to model performance significantly below zero-shot levels, underscoring the importance of careful ICE data preparation and further research to mitigate the negative impact of ICL distribution shifts on generalization.

2. \textbf{How MLLMs leverage ICL to learn reliable knowledge is a crucial question.} Consistent with our findings in IID experiments, Gemini appears to extract more information from ICE compared to GPT-4o. While beneficial under IID ICL conditions, this becomes detrimental when distribution shifts occur, potentially leading to the acquisition of spurious knowledge and severe performance degradation. This raises a critical question: How should MLLMs and LLMs learn and utilize knowledge from ICE, discerning between factual information and statistical biases introduced by small sample sizes? When ICE and test data are from the same distribution, models should maximize knowledge extraction from ICE to efficiently acquire core knowledge. However, when distribution shifts are present, indiscriminate learning from ICE may be counterproductive. This provides a crucial research direction for improving MLLM generalization through ICL.

\section{Conclusion}

We introduce COUNTS, a comprehensive benchmark dataset designed to address the critical gap in evaluating and improving the OOD generalization capabilities of object detectors and MLLMs on fine-grained visual grounding tasks. By providing a large-scale dataset with diverse real-world distribution shifts and fine-grained annotations, COUNTS facilitates the development of more robust and reliable models for complex visual tasks. The proposed benchmarks, O(OD)$^2$ and OODG, enable a thorough investigation of the factors influencing OOD generalization and offer insights for advancing research in this crucial area.
\section*{Acknowledgement}
This work was supported in part by China National Postdoctoral Program for Innovative Talents
(BX20240203), Beijing Municipal Science and Technology Project (No. Z241100004224009), NSFC(No. 62425206, 62141607).

\clearpage
\appendix
\onecolumn
\setcounter{page}{1}
\maketitlesupplementary
\section{More details about COUNTS}

\subsection{Seletion of Domains}
We identified 14 distinct domains: dim, painting, snow, sand, handmade, street, road, water, grass, indoor, mountain, sky, tree, and occlusion. These domains are prevalent in everyday visual experiences and are likely to exhibit distinct correlations with object occurrences. Importantly, while objects within the dataset samples are typically situated in the foreground, the identified domains often function as background elements or overarching image attributes. Consequently, we annotate domain labels at the image level, while providing bounding box annotations at the object level.

These domains are defined and described as follows.
\begin{itemize}
\item \textbf{Dim:} Images captured in suboptimal lighting conditions, often during dusk or dawn, resulting in objects blending with dimly lit backgrounds or obscured textures due to backlighting.

\item \textbf{Painting:} Images where the primary objects are depicted within paintings rather than being real-world objects. This domain introduces potential discrepancies between the depicted objects and their real-world counterparts, as well as differences in background characteristics.

\item \textbf{Snow:} Images featuring prominent snowy landscapes. Objects may be situated against bright, uniform snow backgrounds or partially covered by snow.

\item \textbf{Handmade:} Images where the primary objects are handcrafted, such as plastic or plush toys. These objects may differ in appearance from their real-world counterparts.

\item \textbf{Street:} Images captured in street settings, typically characterized by outdoor, less open environments.

\item \textbf{Road:} Images with road backgrounds, encompassing urban roads and highways. Compared to the "street" domain, "road" scenes predominantly feature major thoroughfares and highways, resulting in more open environments.

\item \textbf{Water:} Images primarily featuring water surfaces or underwater scenes. Objects may be partially submerged or reflected in the water.

\item \textbf{Grass:} Images with grassy backgrounds, primarily covered in low-lying vegetation, often appearing yellow or green. Objects may be partially obscured by vegetation.

\item \textbf{Indoor:} Images captured in indoor environments, such as residences, stadiums, or exhibition halls. These scenes may exhibit specific lighting conditions and object states (e.g., objects displayed in showcases).

\item \textbf{Mountain:} Images taken in mountainous regions, potentially featuring variations in viewing angles and exposed rock formations as backgrounds.

\item \textbf{Sky:} Images with prominent sky backgrounds. Objects may be located at a distance from the camera or captured from a low-angle perspective.

\item \textbf{Tree:} Images featuring trees, forests, or other tall vegetation as backgrounds, often appearing green or dark green.

\item \textbf{Occlusion:} Images where the primary objects are partially or fully occluded, such as in crowded scenes where people are obscured by others.
\end{itemize}

\subsection{Statistics of COUNTS}
We report detailed statistical information on COUNTS\footnote{The COUNTS dataset can be found at https://huggingface.co/datasets/jianshengli/COUNTS} in this section. The number of samples per domain is presented in Table \ref{tab:app-counts-domain} and the number of samples per category is presented in Table \ref{tab:app-counts-cate}.

\begin{table*}[th]
\centering

\resizebox{\linewidth}{!}{
\begin{tabular}{lcccccccccccccc|c}
\toprule

Domain &  Dim & Painting & Snow & Sand& Handmade& Street& Road& Water& Grass& Indoor& Mountain& Sky& Tree& Occlusion & All  \\
\midrule
Images &26181&4167&3321&1788&2150&22474&12637&8915&14620&82719&1452&16771&45573&4215&246983 \\
Objects &130448&14369&15029&15022&7671&185883&96506&34804&81530&470915&7578&74513&314532&28932&1477732 \\

\bottomrule
\end{tabular}}
\caption{Statistics of domains in COUNTS.}
\label{tab:app-counts-domain}
\end{table*}

\begin{table*}[th]
\centering

\resizebox{\linewidth}{!}{
\begin{tabular}{lccccccccccccccc}
\toprule

Category &person&sculpture&building&bicycle&wheel&car&train&glasses&truck&football&dress&dog&drum&guitar&motorcycle
    \\
\midrule
Objects &564546&2378&41036&10170&48787&44418&3129&7952&2794&275&8228&5191&4239&5039&4307 \\
\midrule

Category &sunglasses&suit&boat&bird&flowerpot&houseplant&horse&poster&goggles&food&trousers&chair&shorts&wineglass &bicyclehelmet   \\
\midrule
Objects &4834&19319&1714&4003&2782&3143&3211&2533&1356&13624&5649&59916&3853&3884&2519 \\
\midrule
Category &umbrella&sunhat&helmet&drink&handbag
    \\
\midrule
Objects &4027&923&5128&9034&1029& \\

\bottomrule
\end{tabular}}
\caption{Statistics of categories in COUNTS.}
\label{tab:app-counts-cate}
\end{table*}
\section{More Experiment Details}
\subsection{Evaluated Object Detectors}
In this section, we introduce the evaluated object detectors via our O(OD)$^2$ benchmark. 
YOLO\cite{redmon2016you} is a real-time object detection model known for its speed and efficiency, using a single neural network to predict bounding boxes and class probabilities directly from full images. 
YOLOv9\cite{wang2024yolov9} is the latest iteration of the YOLO series, featuring improved accuracy, speed, and architecture modifications for enhanced object detection performance.
RetinaNet\cite{lin2017focal} is a one-stage object detector that addresses the class imbalance problem in object detection by introducing a focal loss function.
DETR\cite{carion2020end} is an object detection model that leverages transformers and bipartite matching to eliminate the need for hand-designed components like anchor boxes and non-maximum suppression.
DINO\cite{zhang2022dino} is a self-supervised learning method for object detection that utilizes a teacher-student framework and contrastive learning to learn robust object representations without manual labeling.

\subsection{Training Details of Detectors}
For YOLOv9, we employed the official YOLOv9 implementation, while other models were implemented using MMDetection. Unless otherwise specified, all detectors were initialized with pre-trained parameters on ImageNet-1k. All the training is conducted for 60 epochs with an initial learning rate of 0.001, decayed by a factor of 0.1 at epochs 20 and 34. A batch size of 32 and a weight decay of 0.0001 were used. All experiments were performed using 8 NVIDIA RTX 4090 GPUs.

\subsection{Prompts for OODG}
We show our prompt for Visual Grounding without and with ICE in Table \ref{tab:prompt-vg-zero} and \ref{tab:prompt-vg-ice}, respectively.

\begin{table*}[th]
\centering

\begin{tabular}{l}
\toprule

Question: What object is in the red box? <image> \\
Option:
(A) \{category 1\} \\
(B) \{category 2\} \\
(C) \{category 3\} \\
(D) \{category 4\} \\
Please respond with the following format: \\
—BEGIN FORMAT TEMPLATE– \\
Answer Choice: [Your Answer Choice Here] \\
—END FORMAT TEMPLATE– \\
Do not deviate from the above format. Repeat the format template for the answer. \\

\bottomrule
\end{tabular}
\caption{The prompt for Visual Grounding without ICE.}
\label{tab:prompt-vg-zero}
\end{table*}

\begin{table*}[th]
\centering

\begin{tabular}{l}
\toprule

Answer the final question according to the following examples. \\
Examples: \\
Question: What object is in the red box? <image> \\
Option: \\
(A) \{category 1\} \\
(B) \{category 2\} \\
(C) \{category 3\} \\
(D) \{category 4\} \\
Please respond with the following format: \\
—BEGIN FORMAT TEMPLATE– \\
Answer Choice: [Your Answer Choice Here] \\
—END FORMAT TEMPLATE– \\
Do not deviate from the above format. Repeat the format template for the answer. \\
Answer choice: \{ground truth\} \\
\\
$\cdots$ (other examples)\\
\\
Final Question: What object is in the red box? <image> \\
Option: \\
(A) \{category 1\} \\
(B) \{category 2\} \\
(C) \{category 3\} \\
(D) \{category 4\} \\
Please respond with the following format: \\
—BEGIN FORMAT TEMPLATE– \\
Answer Choice: [Your Answer Choice Here] \\
—END FORMAT TEMPLATE– \\
Do not deviate from the above format. Repeat the format template for the answer. \\
\bottomrule
\end{tabular}
\caption{The prompt for Visual Grounding with ICE.}
\label{tab:prompt-vg-ice}
\end{table*}

We show our prompt for Recognition and Localization in Table \ref{tab:prompt-rl}.

\begin{table*}[th]
\centering

\begin{tabular}{l}
\toprule

Knowing that the dimensions of the input image are weight {weight} and height {height}, \\
find all the objects of {category 1} in the image and return the coordinates [X, Y, W, H] \\ 
of the rectangular box in which they are located, \\
where X, Y denote the horizontal coordinate and the vertical coordinate of the \\ 
upper-left corner of the rectangular box, respectively. \\
W, H denote the weight and the height of the rectangular box, respectively. \\
Please respond with the following format: \\
—BEGIN FORMAT TEMPLATE– \\
Box item: [the coordinates of the box] \\
—END FORMAT TEMPLATE– \\
Do not deviate from the above format. Repeat the format template for the answer. \\

\bottomrule
\end{tabular}
\caption{The prompt for Recognition and Localization.}
\label{tab:prompt-rl}
\end{table*}

We show our prompt for Visual and Semantic Mapping in Table \ref{tab:prompt-map}.

\begin{table*}[th]
\centering

\begin{tabular}{l}
\toprule

Question: What object is in the blue box, red box, yellow box, and green box, respectively? <image> \\
Option: \\
(A) \{category 1\} \\
(B) \{category 2\} \\
(C) \{category 3\} \\
(D) \{category 4\} \\
Please respond with the following format: \\
—BEGIN FORMAT TEMPLATE– \\
Answer Choice:  \\
blue box:[Your Answer Choice Here] \\
red box:[Your Answer Choice Here] \\
yellow box:[Your Answer Choice Here] \\
green box:[Your Answer Choice Here] \\
—END FORMAT TEMPLATE– \\
Do not deviate from the above format. Repeat the format template for the answer.
 \\

\bottomrule
\end{tabular}
\caption{The prompt for Visual and Semantic Mapping.}
\label{tab:prompt-map}
\end{table*}
\section{More Experimental Results}

We present the performance of various MLLMs on the OODG benchmark, including GPT-4o, GPT-4-turbo, Gemini-1.5-Flash, Gemini-1.5-Pro, and Claude-3-Opus. Table \ref{tab:app-grounding} and \ref{tab:app-grounding-2} detail the results for each model on the Visual Grounding task. 

While GPT-4-turbo demonstrates slightly weaker zero-shot performance compared to GPT-4o, their performance becomes comparable when distribution shifts exist between ICE and test samples, suggesting minimal influence from ICE on GPT-4-turbo. Overall, Gemini-1.5-Pro exhibits marginally stronger generalization capabilities than Gemini-1.5-Flash, but both are significantly affected by ICL and experience notable performance degradation when distribution shifts are present between ICE and test samples. Claude-3-Opus, while demonstrating slightly weaker zero-shot performance compared to GPT-4 and Gemini-1.5, appears less susceptible to being misled by biased ICE during ICL. These findings highlight significant variations in the generalization capabilities of different MLLMs and their varying degrees of reliance on and susceptibility to ICL. This underscores the importance of further research into leveraging ICL to enhance MLLM generalization while mitigating the potential negative impacts of biased or misaligned ICE.

\begin{table*}[th]
\centering
\caption{Results of Visual Grounding of more models on \textit{sky}, \textit{occlusion}, and \textit{grass}. The selection accuracy is reported. \textit{S}, \textit{M}, and \textit{L} indicate small, medium, and large objects, respectively.}
\resizebox{0.9\linewidth}{!}{
\begin{tabular}{lccccccccccc}
\toprule
\multirow{2}*{Setting} & \multirow{2}*{Model} & \multirow{2}*{ICE}  & \multicolumn{3}{c}{Sky} & \multicolumn{3}{c}{Occlusion} & \multicolumn{3}{c}{Grass}  \\ 
& & & S & M & L & S & M & L & S & M & L \\

\midrule

\multirow{5}*{1} & GPT-4o &  0 & 0.434&0.727 &0.889 &0.362 &0.891 &0.805 &0.969 & 0.939&0.818 \\ 
& GPT-4-turbo-2024-04-09&0 & 0.286 & 0.500&0.917 &0.308 &0.444 &0.833 &0.813 &0.833 &0.667   \\

\cmidrule(lr){2-12}
 & Gemini-1.5-Flash &  0 &0.444 &0.667 & 0.778& 0.355&0.674 &0.628 &0.424 & 0.343&0.615  \\
 & Gemini-1.5-Pro&0&0.438&0.671&0.744&0.372&0.674&0.624&0.428&0.329&0.614\\

\cmidrule(lr){2-12}
 & Claude-3-opus-2024-02-29&0&0.000&0.286&1.000&0.111&0.571&0.222&0.286&0.429&0.429\\
 \cmidrule(lr){2-12}
 & groundingLMM&0&0.333&0.667&1.000&0.387&0.714&0.444&0.667&0.818&0.818\\
 \cmidrule(lr){2-12}
 & Qwen&0&0.286&0.667&0.889&0.333&0.667&0.355&0.424&0.667&0.714\\
\midrule

\multirow{10}*{2} & \multirow{2}*{GPT-4o} &  1&0.445 &0.778 &0.778 &0.571 &1.000 &1.000 &0.857 &0.857 &0.714\\ 
&  & 2 &0.445&0.667 & 0.778&0.143 &0.857 &1.000 &0.857 &0.857 &0.571 \\
& \multirow{2}*{GPT-4-turbo-2024-04-09} & 1 &0.482&0.803&0.819&0.608&1.000&1.000&0.937&0.924&0.772\\
& & 2&0.487&0.679&0.808&0.157&0.891&1.000&0.917&0.870&0.598 \\
\cmidrule(lr){2-12}
 & \multirow{2}*{Gemini-1.5-Flash} &  1 &0.444 &0.714 &1.000 &0.143 &1.000 &0.857 & 0.833&1.000 &0.429 \\ 
&  & 2 &0.444&0.857 &1.000 &0.143 &1.000 &1.000 &1.000 & 1.000& 0.788\\

 & \multirow{2}*{Gemini-1.5-Pro} & 1&0.465&0.777&1.000&0.143&1.000&0.910&0.852&1.000&0.459 \\
 & & 2&0.458&0.862&1.000&0.148&1.000&1.000&1.000&1.000&0.837\\

\cmidrule(lr){2-12}
 & \multirow{2}*{Claude-3-opus-2024-02-29} & 1&0.250&0.562&1.000&0.111&0.222&0.147&0.400&0.083&0.556\\
 & &  2&0.400&0.765&0.846&0.750&0.444&0.333&0.467&0.353&0.308
 \\

\midrule

\multirow{10}*{3} & \multirow{2}*{GPT-4o} &  1 &0.143 &0.357 &0.857 &0.357 &0.786 & 0.929&0.929 & 0.643&0.571 \\ 
&  & 2 &0.429 & 0.556& 0.492&0.385 &1.000 &0.929 &0.786 &0.571 & 0.500\\

& \multirow{2}*{GPT-4-turbo-2024-04-09} & 1 &0.146&0.382&0.909&0.369&0.830&0.940&1.000&0.680&0.597\\
& & 2 &0.432&0.602&0.499&0.421&1.000&0.945&0.845&0.590&0.529\\
\cmidrule(lr){2-12}
 & \multirow{2}*{Gemini-1.5-Flash} &  1 &0.222 &0.143 &0.143 &0.143 &0.143 &0.143 &0.111 &0.111 &0.111  \\ 
&  & 2 &0.000 &0.000 &0.222 &0.143 &0.143 &0.143 &0.133 &0.133 &0.122 \\

 & \multirow{2}*{Gemini-1.5-Pro} & 1 &0.238&0.154&0.151&0.156&0.145&0.148&0.121&0.116&0.111\\
 & & 2 &0.000&0.000&0.237&0.145&0.153&0.148&0.143&0.142&0.130\\

\cmidrule(lr){2-12}
 & \multirow{2}*{Claude-3-opus-2024-02-29} & 1&0.208&1.000&0.636&0.111&0.857&0.571&0.232&0.000&1.000\\
 & &  2&0.222&0.778&0.333&0.429&0.714&0.714&0.571&0.143&1.000\\

\midrule

\multirow{10}*{4} & \multirow{2}*{GPT-4o} &  1 & 0.133 &0.286 &1.000 &0.429 &0.857 &0.857 &0.714 &0.714 &0.571 \\ 
&  & 2 &0.143 &0.286 &1.000 &0.286 &0.571 &0.857 &1.000 &0.857 &0.857 \\

& \multirow{2}*{GPT-4-turbo-2024-04-09} & 1& 0.143&0.310&1.000&0.441&0.887&0.926&0.765&0.751&0.602 \\
& & 2 &0.145&0.288&1.000&0.294&0.617&0.937&1.000&0.891&0.919\\
\cmidrule(lr){2-12}
 & \multirow{2}*{Gemini-1.5-Flash} &  1 & 0.000 &0.111 &0.143 &0.111& 0.133&0.127 &0.111 &0.111 &0.286 \\ 
&  & 2 &0.111 &0.111 &0.143 &0.111 &0.137 &0.111 &0.143 &0.143 &0.143 \\

 & \multirow{2}*{Gemini-1.5-Pro} & 1& 0.000&0.120&0.145&0.112&0.144&0.136&0.117&0.112&0.313\\
 & & 2& 0.111 &0.111 &0.143 &0.111 &0.137 &0.111 &0.143 &0.143 &0.143 \\

\cmidrule(lr){2-12}
 & \multirow{2}*{Claude-3-opus-2024-02-29} & 1&0.111&0.833&0.722&0.208&0.714&0.429&0.208&0.667&1.000\\
 & & 2&0.208&0.611&0.333&0.571&1.000&0.200&0.232&0.143&1.000\\

\midrule

\multirow{10}*{5} & \multirow{2}*{GPT-4o} &  4 &0.286 & 0.500&0.917 &0.308 &0.444 &0.833 &0.813 &0.833 &0.667 \\ 
&  & 8 &0.714 &0.875 &0.833 &0.311 &0.556 &0.833 &0.938 &0.611 &0.583  \\

& \multirow{2}*{GPT-4-turbo-2024-04-09} & 4 &0.294&0.518&0.964&0.318&0.468&0.890&0.853&0.885&0.730\\
& & 8&0.727&0.938&0.846&0.329&0.587&0.857&0.966&0.661&0.614\\
\cmidrule(lr){2-12}
 & \multirow{2}*{Gemini-1.5-Flash} &  4 &0.500 &0.875 & 0.888& 0.534&0.473 &0.500 &0.889 &0.875 &0.500\\ 
&  & 8 &0.467 & 0.889&0.875 & 0.690&0.544 &0.812 &0.844 &0.833 &0.833 \\

 & \multirow{2}*{Gemini-1.5-Pro} & 4&0.546&0.962&0.911&0.559&0.515&0.541&0.969&0.959&0.509\\
 & & 8&0.497&0.937&0.889&0.738&0.573&0.835&0.859&0.852&0.839\\

\cmidrule(lr){2-12}
 & \multirow{2}*{Claude-3-opus-2024-02-29} & 4&0.143 & 0.172&0.917 &0.308 &0.444 &0.833 &0.813 &0.833 &0.667 \\
 & &  8&0.167 & 0.214&0.917 &0.308 &0.444 &0.833 &0.813 &0.833 &0.667 \\

\bottomrule
\end{tabular}}
\label{tab:app-grounding}
\end{table*}

\begin{table*}[th]
\centering
\caption{Results of Visual Grounding of more models on \textit{water}, \textit{dim}, and \textit{handmake}. The selection accuracy is reported. }
\resizebox{\linewidth}{!}{
\begin{tabular}{lccccccccccc|c}
\toprule
\multirow{2}*{Setting} & \multirow{2}*{Model} & \multirow{2}*{ICE}    & \multicolumn{3}{c}{Water} & \multicolumn{3}{c}{Dim}&\multicolumn{3}{c}{Handmake} & \multirow{2}*{Overall} \\ 
  &  &  & S & M & L & S & M & L & S & M & L & \\

\midrule

\multirow{5}*{1} & GPT-4o &  0   & 0.554 &0.797 &0.860 &0.626 & 0.727& 0.737&0.353 &0.595 &0.535&0.701 \\ 
& GPT-4-turbo-2024-04-09&0& 0.375 &0.556 &0.750 &0.333 & 0.588& 0.769&0.133&0.352 &0.462&0.567  \\

\cmidrule(lr){ 2-13}
 & Gemini-1.5-Flash &  0   &0.533 &0.783 &0.858&0.347 & 0.716& 0.570&0.333 &0.608 &0.575&0.570 \\
 & Gemini-1.5-Pro&0& 0.554&0.766&0.883&0.351&0.730&0.593&0.325&0.617&0.578&0.572 \\

\cmidrule(lr){ 2-13}
 & Claude-3-opus-2024-02-29&0 &0.571&0.429&0.714&0.286&0.714&0.857&0.143&0.714&0.714&0.471\\
 \cmidrule(lr){ 2-13}
  & groundingLMM&0&0.554&0.857&1.000&0.429&0.857&0.857&0.333&0.429&0.513&0.648\\
 \cmidrule(lr){2-12}
 & Qwen&0&0.533&0.667&0.750&0.385&0.625&0.714&0.286&0.444&0.555&0.551\\
\midrule

\multirow{10}*{2} & \multirow{2}*{GPT-4o} &  1  &0.143 &0.429 &1.000 &0.333 & 0.778& 0.750&0.000 &0.680 &1.000&0.673\\ 
&  & 2   &0.554 &0.797 &0.860 &0.333 & 0.778& 0.778&0.000 &0.825 &1.000&0.672\\
& \multirow{2}*{GPT-4-turbo-2024-04-09} & 1  &0.149&0.449&1.000&0.350&0.779&0.774&0.000&0.720&1.000&0.710\\
& & 2& 0.562&0.810&0.900&0.338&0.843&0.835&0.000&0.860&1.000&0.699 \\
\cmidrule(lr){ 2-13}
 & \multirow{2}*{Gemini-1.5-Flash} &  1   &0.286 &0.857 & 1.000&0.222 & 1.000 & 0.889&0.000 &0.637 &1.000&0.684\\ 
&  & 2 & 0.143 &0.857 &1.000 &0.333 & 1.000& 0.889&0.000 &0.671 &1.000&0.729 \\

 & \multirow{2}*{Gemini-1.5-Pro} & 1 &0.290&0.916&1.000&0.229&1.000&0.908&0.000&0.682&1.000&0.714 \\
 & & 2 &0.144&0.937&1.000&0.358&1.000&0.920&0.000&0.694&1.000&0.761 \\

\cmidrule(lr){ 2-13}
 & \multirow{2}*{Claude-3-opus-2024-02-29} & 1 &0.231&0.444&0.667&0.232&0.232&0.571&0.400&0.429&0.429&0.387 \\
 & &  2 &0.077&0.333&0.583&0.167&0.214&0.917&0.250&0.333&0.583&0.451
 \\

\midrule

\multirow{10}*{3} & \multirow{2}*{GPT-4o} &  1   &0.143 &0.429 &0.786 &0.222 & 0.722& 0.667&0.000 &0.889 &1.000&0.605 \\ 
&  & 2  &0.143 &0.429 &0.929 &0.278 & 0.778& 0.778&0.000 &0.714 &1.000&0.604 \\

& \multirow{2}*{GPT-4-turbo-2024-04-09} & 1  &0.147&0.450&0.816&0.240&0.780&0.694&0.000&0.916&1.000&0.606 \\
& & 2  &0.145&0.448&0.987&0.299&0.789&0.847&0.000&0.778&1.000&0.639\\
\cmidrule(lr){ 2-13}
 & \multirow{2}*{Gemini-1.5-Flash} &  1  &0.286 &0.714 &0.857 &0.111 & 0.444& 0.444&0.000 &0.571 &1.000&0.322 \\ 
&  & 2   &0.286 &0.714 &0.857 &0.222 & 0.444& 0.444&0.000 & 0.571&1.000&0.328 \\

 & \multirow{2}*{Gemini-1.5-Pro} & 1 & 0.286&0.773&0.858&0.116&0.461&0.452&0.000&0.593&1.000&0.327\\
 & & 2  &0.290&0.741&0.869&0.222&0.453&0.481&0.000&0.596&1.000&0.350\\

\cmidrule(lr){ 2-13}
 & \multirow{2}*{Claude-3-opus-2024-02-29} & 1 &0.222&0.429&0.857&0.111&0.000&0.857&0.429&0.208&1.000&0.485\\
 & &  2 &0.143&0.143&0.714&0.000&0.200&0.857&0.000&0.222&0.857&0.447\\

\midrule

\multirow{10}*{4} & \multirow{2}*{GPT-4o} &  1   &0.143 &0.429 &0.857 &0.000 & 0.556& 0.667& 0.342& 0.432&1.000&0.580 \\ 
&  & 2   &0.143 &0.429 &0.857 &0.111 & 0.667& 0.778&0.000 &0.424 &1.000&0.595\\

& \multirow{2}*{GPT-4-turbo-2024-04-09} & 1 &0.149&0.445&0.903&0.000&0.572&0.706&0.375&0.441&1.000&0.584  \\
& & 2  &0.150&0.449&0.921&0.115&0.698&0.832&0.000&0.433&1.000&0.596\\
\cmidrule(lr){ 2-13}
 & \multirow{2}*{Gemini-1.5-Flash} &  1   &0.143 &0.857 &0.857 &0.111 & 0.222& 0.222&0.000 &0.714 &1.000&0.309\\ 
&  & 2   &0.143 &1.000 &0.714 &0.000 & 0.222& 0.222&0.000 &0.424 &1.000&0.280\\

 & \multirow{2}*{Gemini-1.5-Pro} & 1 &0.145&0.932&0.893&0.114&0.237&0.234&0.000&0.734&1.000&0.305 \\
 & & 2  &0.143 &1.000 &0.714 &0.000 & 0.222& 0.222&0.000 &0.424 &1.000&0.280 \\

\cmidrule(lr){ 2-13}
 & \multirow{2}*{Claude-3-opus-2024-02-29} & 1 &0.111&0.429&0.714&0.232&0.222&0.929&0.000&0.111&0.857&0.472 \\
 & & 2 &0.333&0.308&0.786&0.286&0.143&0.857&0.273&0.400&1.000&0.482 \\

\midrule

\multirow{10}*{5} & \multirow{2}*{GPT-4o} &  4   &0.375 &0.556 &0.750 &0.333 & 0.588& 0.769&0.133&0.352 &0.462&0.567\\ 
&  & 8   &0.313 &0.778 &0.750 &0.143 & 0.800& 0.923&0.266 &0.412 &0.625&0.621 \\

& \multirow{2}*{GPT-4-turbo-2024-04-09} & 4  &0.380&0.606&0.761&0.340&0.643&0.811&0.140&0.374&0.480&0.581 \\
& & 8 &0.336&0.800&0.759&0.147&0.826&0.997&0.277&0.445&0.656&0.654 \\
\cmidrule(lr){ 2-13}
 & \multirow{2}*{Gemini-1.5-Flash} &  4   &0.375 &0.556 &0.750 &0.347 & 0.716& 0.570&0.400 &0.313 &0.769&0.608\\ 
&  & 8   &0.313 &0.778 &0.750 &0.347 & 0.716& 0.570&0.467 &0.353 &0.462&0.652\\

 & \multirow{2}*{Gemini-1.5-Pro} & 4 &0.409&0.557&0.806&0.373&0.779&0.575&0.403&0.325&0.796&0.639 \\
 & & 8 &0.319&0.848&0.818&0.357&0.739&0.591&0.478&0.373&0.464&0.667 \\

\cmidrule(lr){ 2-13}
 & \multirow{2}*{Claude-3-opus-2024-02-29} & 4  &0.375 &0.556 &0.750 &0.333 & 0.588& 0.769&0.133&0.352 &0.462& 0.525\\
 & &  8  &0.375 &0.556 &0.750 &0.333 & 0.588& 0.769&0.133&0.352 &0.462& 0.529\\

\bottomrule
\end{tabular}}
\label{tab:app-grounding-2}
\end{table*}

\subsection{Visual and Semantic Mapping}
We introduce a more challenging experimental setting, termed Visual and Semantic Mapping, which aims to assess a model's ability to align visual regions with corresponding semantic information. Specifically, the model is presented with a set of visual regions and semantic descriptions, and tasked with establishing cross-modal correspondences between these elements. For simplicity, we formulate this task as a ranking problem, wherein the model observes visual regions in a fixed order and is then asked to rank a sequence of semantic descriptions to match the observed visual order. The prompt format for this task is detailed in Table \ref{tab:app-mapping}.

The findings from the Visual and Semantic Mapping experiments align with those from the Visual Grounding and Recognition and Localization tasks, namely that GPT-4 exhibits marginally stronger zero-shot grounding capabilities compared to Gemini. Furthermore, Gemini demonstrates a significantly higher degree of reliance on and utilization of ICE than GPT-4. Consequently, Gemini's performance improves notably when presented with IID ICE, but its generalization and stability are more susceptible to adverse effects when biased ICE are encountered.

\begin{table*}[th]
\centering
\caption{Results of Visual and Semantic Mapping. Mapping accuracy is reported. }
\resizebox{\linewidth}{!}{
\begin{tabular}{lccccccccccc|c}
\toprule
\multirow{2}*{Setting} & \multirow{2}*{Model} & \multirow{2}*{ICE}  & \multicolumn{3}{c}{Occlusion} & \multicolumn{3}{c}{Dim} & \multicolumn{3}{c}{Handmake} & \multirow{2}*{Overall} \\ 
& & & S & M & L & S & M & L & S & M & L   \\
\midrule
\multirow{2}*{1} & GPT-4o &  0&0.390&0.978&0.831&0.682&0.790&0.765&0.376&0.615&0.577&0.667  \\ 
\cmidrule(lr){2-13}
 & Gemini-1.5-Flash &  0& 0.382&0.735&0.670&0.361&0.765&0.608&0.356&0.644&0.610&0.570 \\ 
\midrule

\multirow{4}*{2} & \multirow{2}*{GPT-4o} &  1&0.620&1.000&1.000&0.343&0.814&0.764&0.000&0.726&1.000&0.707  \\ 
&  & 2&0.151&0.889&1.000&0.361&0.812&0.814&0.000&0.902&1.000&0.660 \\
\cmidrule(lr){2-13}
 & \multirow{2}*{Gemini-1.5-Flash} &  1&0.154&1.000&0.859&0.244&1.000&0.971&0.000&0.698&1.000&0.668  \\ 
&  & 2 &0.132&1.000&1.000&0.343&1.000&0.832&0.000&0.618&1.000&0.664 \\

\midrule

\multirow{4}*{3} & \multirow{2}*{GPT-4o} &  1 &0.358&0.774&0.979&0.204&0.670&0.733&0.000&0.908&1.000&0.625 \\ 
&  & 2 &0.412&1.000&0.880&0.301&0.764&0.783&0.000&0.700&1.000&0.649 \\
\cmidrule(lr){2-13}
 & \multirow{2}*{Gemini-1.5-Flash} &  1&0.130&0.131&0.155&0.122&0.450&0.418&0.000&0.547&1.000&0.328  \\ 
&  & 2 &0.151&0.131&0.146&0.230&0.467&0.458&0.000&0.601&1.000&0.354 \\

\midrule

\multirow{4}*{4} & \multirow{2}*{GPT-4o} &  1 &0.414&0.877&0.849&0.000&0.526&0.654&0.345&0.428&1.000&0.566 \\ 
&  & 2 &0.307&0.551&0.919&0.116&0.607&0.856&0.000&0.388&1.000&0.527 \\
\cmidrule(lr){2-13}
 & \multirow{2}*{Gemini-1.5-Flash} &  1 &0.101&0.143&0.118&0.106&0.237&0.221&0.000&0.718&1.000&0.294 \\ 
&  & 2&0.100&0.127&0.101&0.000&0.227&0.218&0.000&0.417&1.000&0.243  \\

\midrule

\multirow{4}*{5} & \multirow{2}*{GPT-4o} &  4&0.295&0.477&0.901&0.319&0.649&0.816&0.131&0.351&0.464&0.489  \\ 
&  & 8&0.346&0.569&0.826&0.152&0.745&0.963&0.288&0.476&0.603&0.552  \\
\cmidrule(lr){2-13}
 & \multirow{2}*{Gemini-1.5-Flash} &  4 &0.565&0.428&0.506&0.328&0.777&0.623&0.437&0.330&0.800&0.533 \\ 
&  & 8&0.653&0.552&0.749&0.369&0.689&0.552&0.431&0.350&0.490&0.537  \\

\bottomrule
\end{tabular}}
\label{tab:app-mapping}
\end{table*}

\subsection{Results of Visual Grounding with more ICE}

In the main paper, we present the capabilities of MLLMs with 0, 1, and 2 in-context examples (ICE). Here, we further analyze their performance when provided with a larger number of ICE (e.g., 4 and 8). The results are shown in Table \ref{tab:app-ice}. 
The results indicate that increasing the number of ICE does not significantly enhance the generalization capabilities of MLLMs. The presence of more biased ICE does not appear to further degrade their performance.

\begin{table*}[th]
\centering
\caption{Results of Visual Grounding with more ICE. Selection accuracy is reported. }
\resizebox{\linewidth}{!}{
\begin{tabular}{lccccccccccc|c}
\toprule
\multirow{2}*{Setting} & \multirow{2}*{Model} & \multirow{2}*{ICE}  & \multicolumn{3}{c}{Occlusion} & \multicolumn{3}{c}{Dim} & \multicolumn{3}{c}{Handmake} & \multirow{2}*{Overall} \\ 
& & & S & M & L & S & M & L & S & M & L   \\

\midrule

\multirow{4}*{2} & \multirow{2}*{GPT-4o} &  4&0.423 &0.571 &1.000 &0.333 &0.778 &0.750 & 0.000 &0.750 &1.000 & 0.623\\ 
&  & 8&0.423 &0.423 &1.000 & 0.333 &0.778 &0.778 &0.000 &0.825 &1.000 & 0.618\\
\cmidrule(lr){2-13}
 & \multirow{2}*{Gemini-1.5-Flash} &  4&0.375 &0.857 &0.857 &0.333 &0.500 &0.750 & 0.000 &0.750 &1.000 & 0.602\\ 
&  & 8&0.400 &0.571 &0.714 &0.333 &0.571 &0.769 &0.000 &0.675 &1.000 & 0.559 \\

\midrule

\multirow{4}*{3} & \multirow{2}*{GPT-4o} &  4&0.385 &0.714 &1.000 &0.222 &0.722 &0.667 & 0.000 &0.875 &1.000 & 0.621\\ 
&  & 8&0.571 &0.827 &0.936 &0.222 &0.722 &0.667 & 0.000 &0.750 &1.000 & 0.633 \\
\cmidrule(lr){2-13}
 & \multirow{2}*{Gemini-1.5-Flash} &  4&0.111 &0.147 &0.147 &0.111 &0.444 &0.444 & 0.000 &0.750 &1.000 & 0.350\\ 
&  & 8 &0.111 &0.147 &0.147 &0.111 &0.444 &0.444 & 0.000 &0.500 &1.000 & 0.323\\

\midrule

\multirow{4}*{4} & \multirow{2}*{GPT-4o} &  4&0.514 &0.857 &0.857 &0.000 &0.667 &0.667& 0.133 &0.350 &0.429 & 0.497\\ 
&  & 8&0.525 &0.936 &0.750 & 0.000 &0.667 &0.667 & 0.133 &0.350 &0.500 & 0.503\\
\cmidrule(lr){2-13}
 & \multirow{2}*{Gemini-1.5-Flash} &  4&0.111 &0.375 &0.200 &0.111 &0.222 &0.167 &0.400 &0.429 &0.429 & 0.272 \\ 
&  & 8&0.111 &0.375 &0.144 &0.111 &0.111 &0.222 &0.400 &0.429 &0.429 &  0.259\\

\bottomrule
\end{tabular}}
\label{tab:app-ice}
\end{table*}

\begin{table*}[t]
\centering
\caption{Results of Recognition and Localization with more ICE. mAP is reported.}
\resizebox{\linewidth}{!}{
\begin{tabular}{lccccccccccc|c}
\toprule
\multirow{2}*{Setting} & \multirow{2}*{Model} & \multirow{2}*{ICE}  & \multicolumn{3}{c}{Occlusion} & \multicolumn{3}{c}{Dim} & \multicolumn{3}{c}{Handmake} & \multirow{2}*{Overall} \\ 
& & & S & M & L & S & M & L & S & M & L   \\

\midrule

\multirow{4}*{2} & \multirow{2}*{GPT-4o} &  4&0.101&0.062&0.088&0.064&0.098&0.084&0.082&0.063&0.092&0.077\\
& &8&0.091&0.045&0.098&0.065&0.089&0.067&0.097&0.056&0.096&0.082\\
\cmidrule(lr){2-13}
 & \multirow{2}*{Gemini-1.5-Flash} &  4&0.098&0.050&0.092&0.062&0.098&0.081&0.054&0.035&0.082&0.075\\ 
&  & 8&0.084&0.025&0.078&0.067&0.080&0.080&0.082&0.051&0.069&0.075 \\

\midrule

\multirow{4}*{3} & \multirow{2}*{GPT-4o} &  4&0.027&0.000&0.019&0.000&0.000&0.043&0.000&0.009&0.016&0.013\\ 
&  & 8&0.023&0.004&0.014&0.000&0.000&0.045&0.000&0.005&0.023&0.013 \\
\cmidrule(lr){2-13}
 & \multirow{2}*{Gemini-1.5-Flash} &  4 &0.000 &0.000 &0.000&0.000 &0.000&0.000&0.000 &0.000 &0.000&0.000 \\ 
&  & 8 &0.000 &0.000 &0.000&0.000 &0.000&0.000&0.000 &0.000 &0.000&0.000\\

\midrule

\multirow{4}*{4} & \multirow{2}*{GPT-4o} &  4&0.009&0.000&0.085&0.001&0.000&0.000&0.091&0.005&0.084&0.031\\ 
&  & 8&0.021&0.004&0.018&0.006&0.000&0.006&0.033&0.050&0.020&0.010\\
\cmidrule(lr){2-13}
 & \multirow{2}*{Gemini-1.5-Flash} &  4&0.104&0.060&0.012&0.003&0.009&0.000&0.050&0.024&0.004&0.022 \\ 
&  & 8&0.002&0.099&0.013&0.008&0.034&0.002&0.098&0.019&0.098&0.045\\

\bottomrule
\end{tabular}}
\label{tab:app-ice}
\end{table*}

\subsection{Comparison of Object Detectors in OOD and i.i.d. Scenarios}

\begin{figure*}[]
    \setlength{\leftskip}{0.3cm}
    \includegraphics[width=1\linewidth,bb=0 0 576 360]{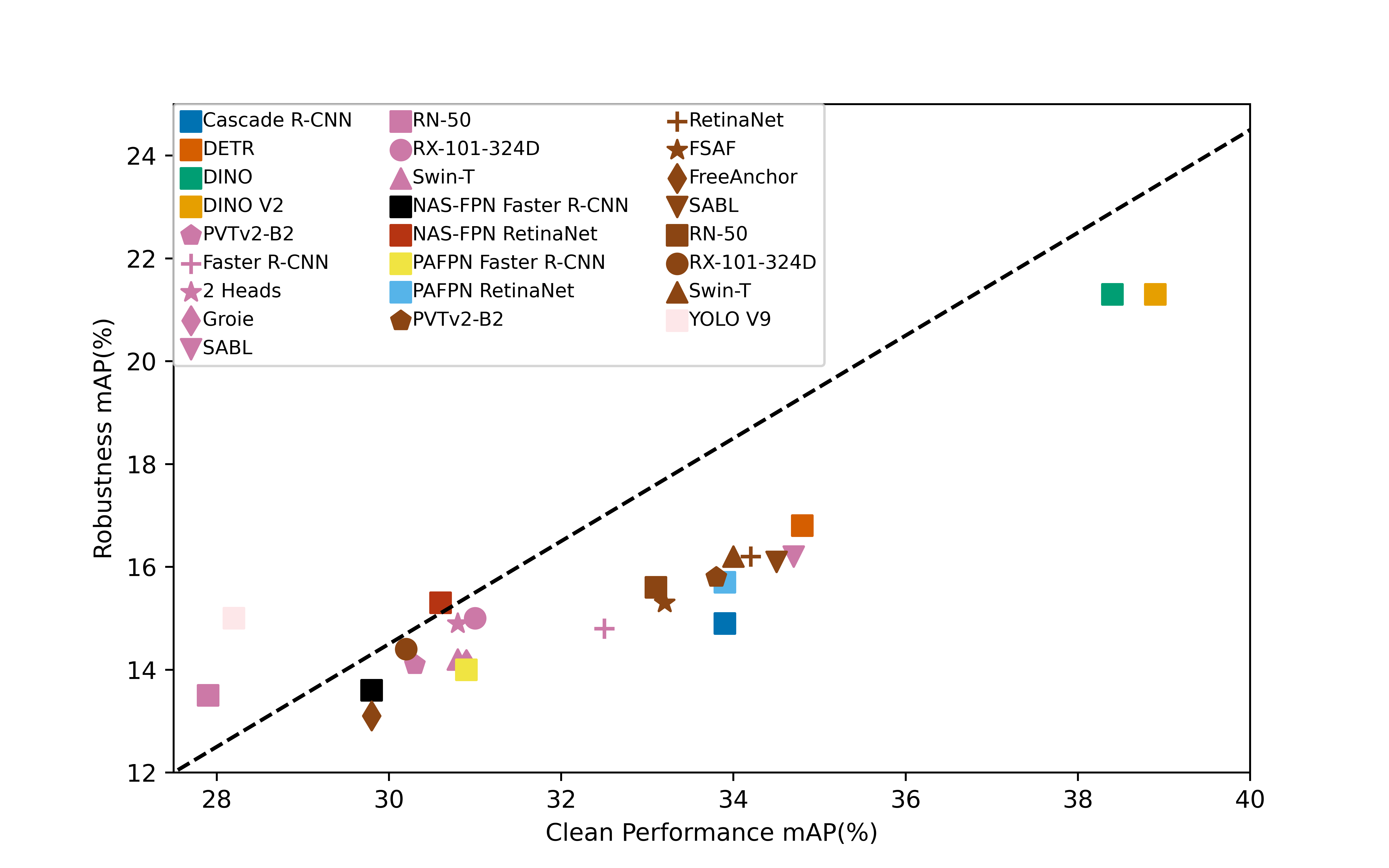}
    \caption{Comparison of current object detectors in OOD and i.i.d. scenarios.}
    \label{fig:map}

\end{figure*}

We summarize the results of existing detectors on COUNT, as shown in Figure \ref{fig:map}. We find that the target detection accuracy of existing detectors in OOD scenarios is significantly lower than that in i.i.d. scenarios, indicating a large room for improvement in the OOD generalization ability of existing detectors.

\subsection{Showcases in OODG}
\begin{figure*}[]
    \setlength{\leftskip}{-1.5cm}
    \vspace{-50pt}
    \includegraphics[width=1.2\linewidth, bb=0 0 540 581.64]{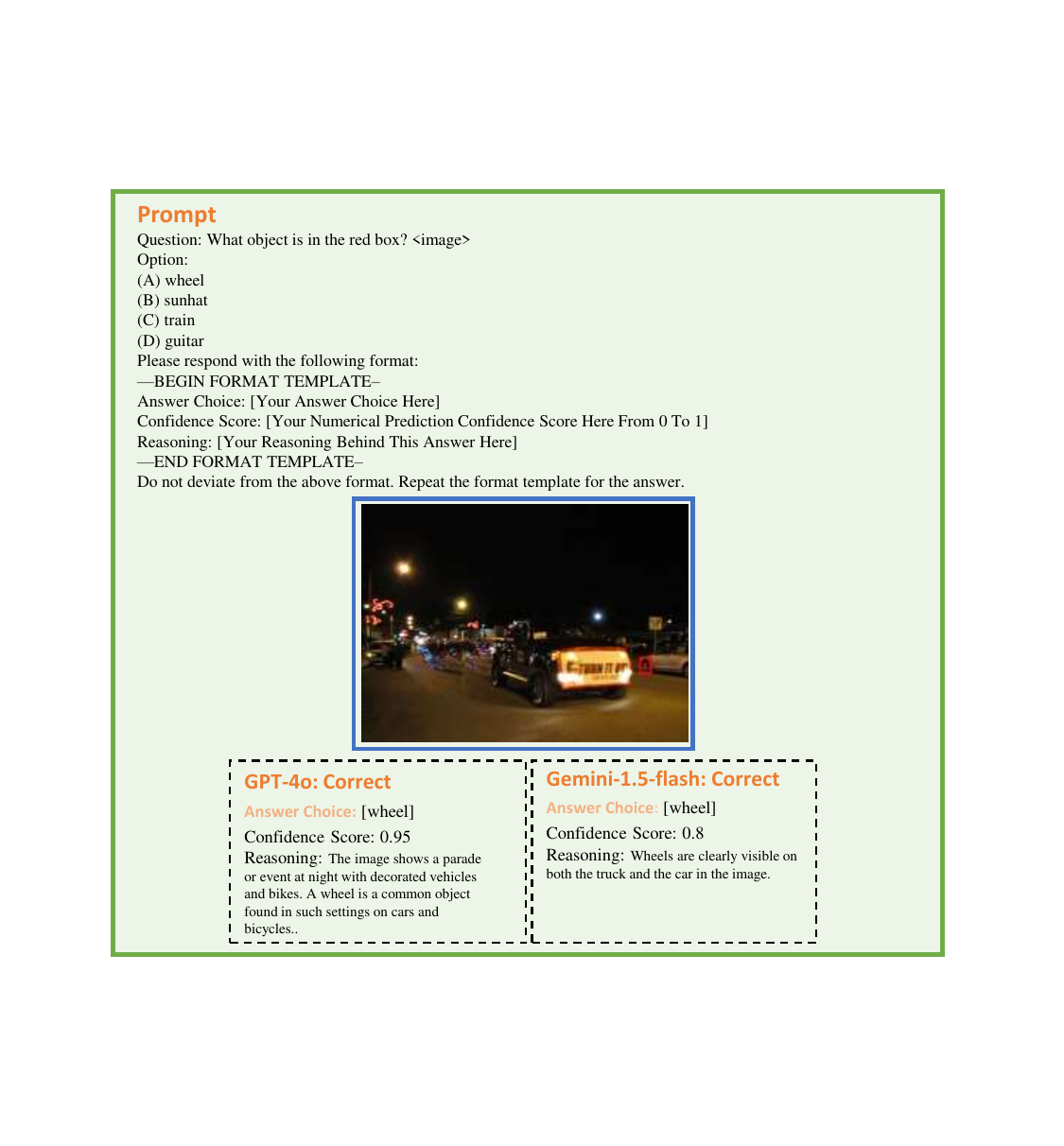}
     \vspace{-80pt}
    \caption{Example of Visual Grounding.}
    \label{fig:showcase1}

\end{figure*}

\begin{figure*}[]
    \setlength{\leftskip}{-1.5cm}
    \vspace{-50pt}
    \includegraphics[width=1.2\linewidth, bb=0 0 540 581.64]{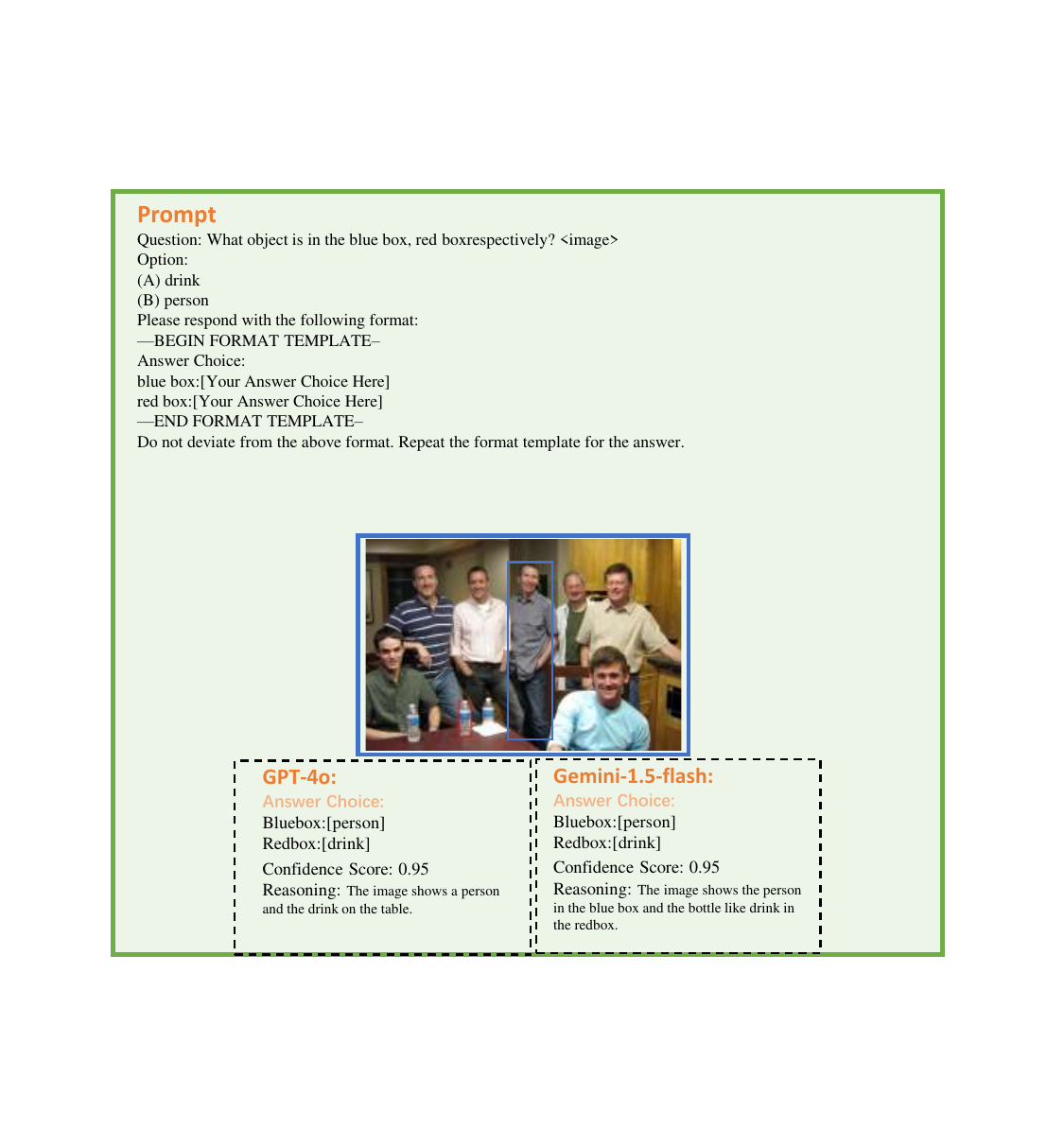}
    \vspace{-80pt}
    \caption{Example of Recognition and Localization.}
    \label{fig:showcase2}
    
\end{figure*}

\begin{figure*}[]
    \setlength{\leftskip}{-1.5cm}
    \vspace{-50pt}
    \includegraphics[width=1.2\linewidth,bb=0 0 540 581.64]{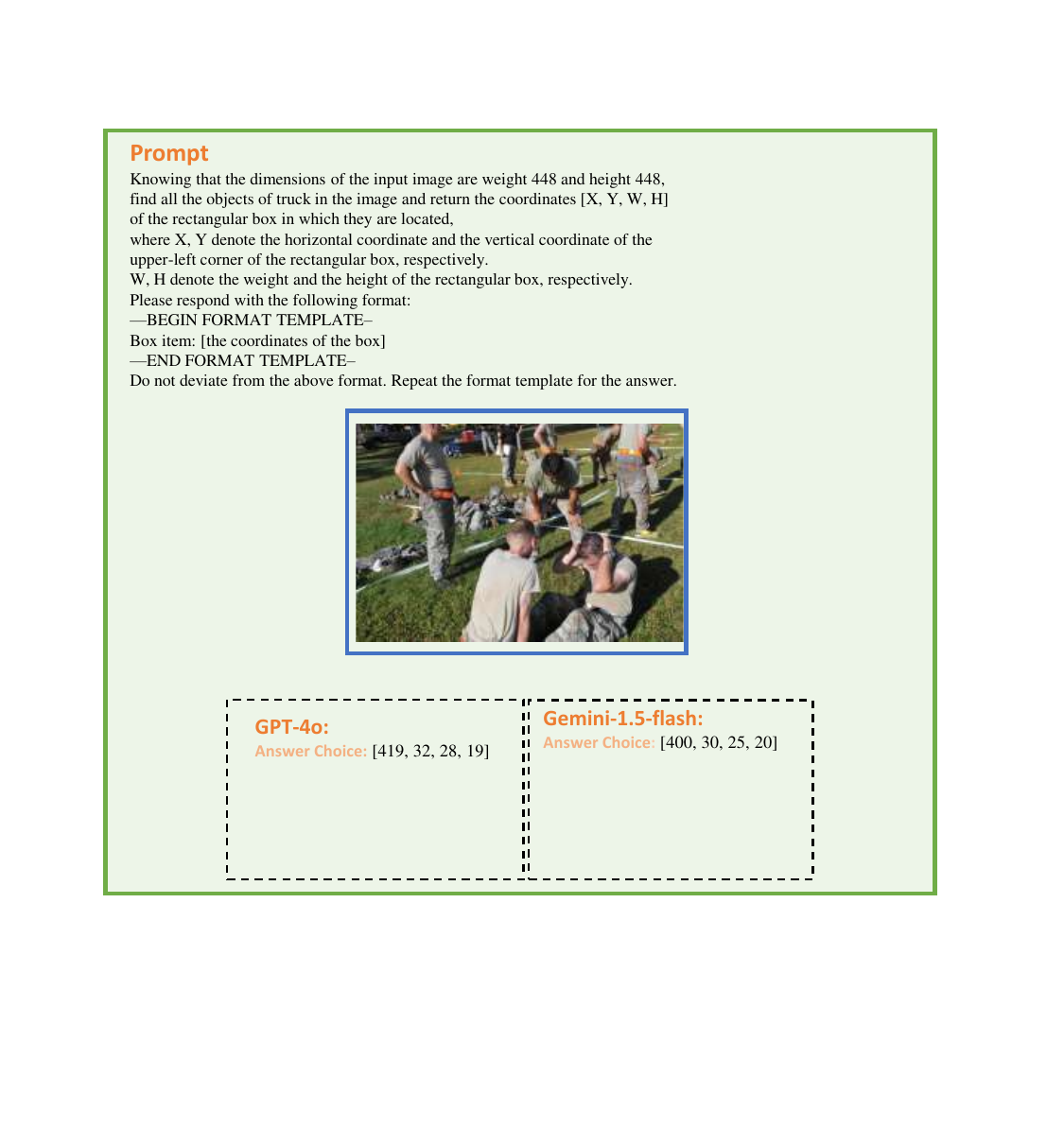}
    \vspace{-110pt}
    \caption{Example of Visual and Semantic Mapping.}
    \label{fig:showcase3}
    \vspace{-50pt}
\end{figure*}

To provide a more intuitive understanding, we present the showcases of Visual Grounding, Recognition and Localization, and Visual and Semantic Mapping in Figure \ref{fig:showcase1}, \ref{fig:showcase2}, and \ref{fig:showcase3}, respectively.

\clearpage

\clearpage

\clearpage

\clearpage
\section{Limitations}
The proposed OODG benchmark presented in this work focuses on evaluating the OOD generalization capabilities of MLLMs with respect to the distribution shift between ICE and test samples. However, it does not explicitly address potential distribution shifts that may arise between the supervised fine-tuning (SFT) data and test samples. Future work could leverage COUNTS to conduct SFT on MLLMs and investigate the impact of distribution shifts occurring during the SFT phase, as well as potential mitigation strategies.

\section{Broader Impact}
The broader impact of this work is significant for both the research community and real-world applications. COUNTS, the introduced dataset, provides a valuable resource for advancing research in OOD generalization, a critical area for deploying robust and reliable machine learning models in real-world scenarios. By focusing on fine-grained object detection and grounding tasks, COUNTS enables the development of models that can accurately perceive and understand visual information even in novel and challenging environments.

Furthermore, the proposed benchmarks, O(OD)$^2$  and OODG, establish standardized evaluation protocols for assessing the OOD generalization capabilities of object detectors and MLLMs, respectively. These benchmarks will serve as valuable tools for researchers and practitioners to gauge the robustness and trustworthiness of their models, ultimately leading to the development of more reliable and adaptable AI systems.

The insights gained from this research can have far-reaching implications across various domains, including autonomous vehicles, robotics, surveillance, and content moderation, where the ability to accurately detect and ground objects under diverse and unpredictable conditions is paramount. Ultimately, this work contributes to the development of AI systems that are more aligned with human visual perception and understanding, thereby increasing their potential for safe and beneficial deployment in real-world applications.

\clearpage
 \small \bibliographystyle{ieeenat_fullname} \bibliography{main}


\end{document}